\documentclass[10pt,journal,cspaper,compsoc]{IEEEtran}

\usepackage{cite}
\usepackage[cmex10]{amsmath}
\usepackage{algorithm}
\usepackage{algorithmic}
\usepackage{array}
\usepackage{url}
\usepackage{stfloats}
\usepackage{graphicx}
\usepackage{color}
\usepackage{eqparbox}
\usepackage{times}
\usepackage{multirow}
\usepackage{epstopdf}
\usepackage{booktabs}
\usepackage{comment}
\usepackage{amssymb}
\usepackage{bigstrut}
\usepackage{multirow}
\usepackage{diagbox}
\usepackage{balance}
\usepackage[table]{xcolor}
\usepackage{caption}
\begin{document}

\title{Dense 3D Face Correspondence}

\author{Syed Zulqarnain Gilani, Ajmal~Mian,
        Faisal~Shafait, and~Ian~Reid 
\IEEEcompsocitemizethanks{\IEEEcompsocthanksitem S.Z. Gilani, A.Mian and F.Shafait are with the School of Computer Science and Software Engineering, The University of Western Australia, 35 Stirling Highway, Crawley, Western Australia, 6009.\protect\\
E-mail: syedzulqarnain.gilani@research.uwa.edu.au} 
\thanks{}
\vspace{-5mm}
\IEEEcompsocitemizethanks{\IEEEcompsocthanksitem I.Reid is with the School of Computer Science 
University of Adelaide, Ingkarni Wardli, North Terrace Campus, Adelaide, SA, Australia \protect\\
} \\
\thanks{}
}

\markboth{IEEE TRANSACTIONS ON PATTERN ANALYSIS AND MACHINE INTELLIGENCE} 
{Shell \MakeLowercase{\textit{et al.}}: Bare Demo of IEEEtran.cls for Computer Society Journals}
%

\IEEEcompsoctitleabstractindextext{%
\begin{abstract}
\textbf{This is the author version of the paper}: Syed Zulqarnain Gilani, Ajmal Mian, Faisal Shafait, and Ian Reid. "Dense 3D face correspondence." IEEE Transactions on Pattern Analysis and Machine Intelligence (TPAMI),40(7), 2017.

We present an algorithm that automatically establishes dense correspondences between a large number of 3D faces. Starting from automatically detected sparse correspondences on the outer boundary of 3D faces, the algorithm triangulates existing correspondences and expands them iteratively by matching points of distinctive surface curvature along the triangle edges. After exhausting keypoint matches, further correspondences are established by generating evenly distributed points within triangles by evolving level set geodesic curves from the centroids of large triangles. A deformable model (K3DM) is constructed from the dense corresponded faces and an algorithm is proposed for morphing the K3DM to fit unseen faces. This algorithm iterates between rigid alignment of an unseen face followed by regularized morphing of the deformable model. We have extensively evaluated the proposed algorithms on synthetic data and real 3D faces from the FRGCv2, Bosphorus, BU3DFE and UND Ear databases using quantitative and qualitative benchmarks. Our algorithm achieved dense correspondences with a mean localisation  error of $1.28$mm on synthetic faces and detected $14$ anthropometric landmarks on unseen real faces from the FRGCv2 database with $3$mm precision. Furthermore, our deformable model fitting algorithm achieved  $98.5\%$ face recognition accuracy on the FRGCv2 and $98.6\%$ on Bosphorus database. Our dense model is also able to generalize to unseen datasets.

\vspace{-3mm}
\end{abstract}

\begin{keywords}
\vspace{-2mm}
Dense correspondence, 3D face, morphing, keypoint detection, level sets, geodesic curves, deformable model.
\end{keywords}}

\maketitle

\vspace{-10mm}
\IEEEdisplaynotcompsoctitleabstractindextext

 \ifCLASSOPTIONpeerreview
 \begin{center} \bfseries EDICS Category: 3-BBND \end{center}
 \fi
%
\IEEEpeerreviewmaketitle


  \noindent\raisebox{2\baselineskip}[0pt][0pt]%
  {\parbox{\columnwidth}{\section{Introduction}\label{sec:introduction}%
  \global\everypar=\everypar}}%
  \vspace{-1\baselineskip}\vspace{-\parskip}\par

\IEEEPARstart{O}{ne} of the canonical tasks  in shape analysis is to find a meaningful mapping between two or more shapes~\cite{van2011}. The process, called shape correspondence, is a pre-requisite for many computer vision, computer graphics and medical image analysis applications. The requisite density of correspondence is often dictated by the underlying shape and target application. Sometimes, sparse correspondence is sufficient to infer shape semantics by matching representative points, for example the four corners of a rectangle or emblematic points on key joints of a human body. However, sparse correspondence is often inadequate in case of articulated shapes~\cite{jain2007,chang2008} where parts of the shape can bend independently or in the correspondence of anatomical shapes which can deform in an elastic manner~\cite{zhang2008}. In such circumstances, dense correspondence is required to guarantee representation of global shape changes, for instance in case of morphing or attribute transfer. Furthermore, very subtle changes within a class of shapes can be detected only if the correspondence between these shapes is dense~\cite{hammond2007}.

In this paper we are concerned with the task of finding dense correspondences between a very large number of similar shapes; in our case 3D scans of human faces.  We do so because this further enables us to generate highly accurate 3D morphable models that can be used
for information transfer between the training set and a test face or between two test faces by morphing the 3D model to fit the test face(s). For example, given the location of anthropometric landmarks~\cite{farkas1994} on the 3D morphable model, these landmarks can be automatically localized on previously unseen test faces~\cite{nair2009}.  Furthermore, dense correspondences and morphable models can be used for 3D face recognition \cite{blanz2003,gilani2017}. Other applications include facial morphometric measurements such as gender scoring\cite{gilani2014c} and asymmetry for syndrome diagnosis~\cite{hammond2007}, statistical shape modelling~\cite{davies2002,heimann2009}, shape interpolation~\cite{alexa2002}, non-rigid shape registration~\cite{aiger2008,brown2007, chang2008}, deformation analysis~\cite{mirzaalian2009} and recognition~\cite{funkhouser2006,passalis2011,prabhu2011}.

While it is possible to manually annotate a small number ($\sim$30) of correspondences for a few 3D faces, it is not feasible to manually identify dense correspondences ($\sim$6,000) between hundreds of 3D faces. The literature also proposes computing dense correspondence by extending manually annotated sparse ones~\cite{kraevoy2004,blanz2003}. However, with the advent of huge 3D face databases like the Facebase Consortium~\cite{hochheiser2011} or Raine dataset~\cite{whitehouse2015,gilani2015b}, this strategy too has become impractical and calls for fully automatic algorithms. Automatically establishing dense correspondences between the 3D faces of two different persons is an extremely challenging task because the facial shape varies significantly amongst individuals depending on their identity, gender, ethnicity and age~\cite{farkas1994} as well as their facial expression and pose. The problem of dense 3D point-to-point correspondences can be formulated as follows. Given a set of $N$ 3D faces, $\mathbf{F}_j=[x_p, y_p,z_p]^T, j=1,\ldots,N, p=1,\ldots,P_j$, the aim is to establish a dense bijective mapping $f:\mathbf{F}_i\rightarrow \mathbf{F}_j (i\neq j)$ over $k$ vertices where $1<<k<min(P_i,P_j)$. Correspondences should cover all regions of the face for high fidelity and should follow the same triangulation for shape consistency.

Existing dense correspondence techniques have one or more of the following limitations: (1) They need manually annotated landmarks on 3D faces for initialization. (2) They use image texture matching to find 3D shape correspondence. (3) They correspond all faces to a single reference face neglecting the global proximity of the 3D faces. (4) They have not been tested on {\em complete} benchmark databases such as the FRGCv2~\cite{phillips2005} or Bosphorus~\cite{savran2008} datasets for face recognition and landmark identification. (5) They have no explicit mechanism of updating the dense correspondence model.

In this context, we propose a fully automatic algorithm for establishing dense correspondences simultaneously between a large number of 3D faces. Our algorithm does not require any manual intervention and relies solely on 3D shape matching to encode accurate facial morphology. We organize the 3D faces into a minimum spanning tree based on bending energy required to deform one shape into the other so that correspondences can be propagated in a reliable way. We propose a mechanism for automatic initialization of a sparse set of correspondences on the outer boundary of the 3D faces.  We  form a triangulation of these correspondences, and iteratively add to the set of points by matching points of distinctive surface curvature along (and close to) the triangulated edges.  After exhausting the possibilities for such matches, we further expand the set of matches by generating points distributed evenly within triangles by evolving level set geodesic curves from the centroids of large triangles. The outcome of our algorithm is a Keypoint-based 3D Deformable Model (K3DM). 

Our second major contribution is a deformable model fitting algorithm where K3DM is used to morph into unseen query faces. Starting from the mean face, the fitting algorithm iterates between two steps. The query face is transformed rigidly to align with the model and the model is deformed using regularized least squares to fit the query face. This algorithm converges in a few iterations and is robust to noise, outlier points, missing points, pose and expression variations.

Our final contribution is an algorithm for augmenting the K3DM. Given the K3DM and a new batch of $M$ faces, we construct a minimum spanning tree using the nearest face to the K3DM as the root node. The K3DM is augmented by adding one face at a time, starting with the root node, and each time updating the model and deforming the updated model to better fit the next face in the spanning tree.

Evaluating dense correspondence techniques is challenging due to  the inherent difficulty of obtaining ground-truth data. In the existing literature, evaluations have mostly been performed on a sparse set of anthropometric facial landmarks~\cite{perakis2013,creusot2013,gilani2015} since these can be manually labelled. However, evaluation on only a few ($\le 20$) anthropometric points does not show how well dense correspondences have generalized to the whole face. Thus, subjective evaluations are frequently performed~\cite{blanz1999} by visually inspecting the quality of morphing between faces~\cite{zhang2008,sun2003}. In this paper, we show how synthetic 3D faces (Facegen\textsuperscript{TM} Modeller) can be used to quantitatively evaluate dense correspondences on a large set of points ($\ge 1,000$). Using the presented deformable face model, we perform extensive experiments for landmark localisation  (Section~\ref{sec:eval-landmark}) 
and face recognition (Section~\ref{sec:eval-face}) using real faces from the FRGCv2~\cite{phillips2005} and BU3DFE~\cite{yin2006} databases. Results show that our algorithm outperforms state-of-the-art application-specific algorithms in each of these areas.

\vspace{-3mm}
\section{{Related Work}}
\vspace{-1mm}
{Existing 3D correspondence techniques can be grouped into descriptor based, model based and optimization based~\cite{van2011}.}

\noindent {\textbf{Descriptor based techniques:} These techniques match local 3D point signatures derived from the curvatures, shape index and normals. However, they are often highly sensitive to surface noise and sampling density~\cite{wang2007} of the underlying geometry~\cite{novatnack2008}. More significantly for our purpose, the density of corresponding points is typically low resulting in correspondences between a very sparse set of anthropometric landmarks.}
	
{One of the earliest works, in this category, for establishing dense correspondence was proposed by Sun and Abidi~\cite{sun2001,sun2003} who projected geodesic contours around a 3D facial point onto their tangential plane and used them as features to match two surfaces. The approach, with minor modifications, was employed by Salazar et al.~\cite{salazar2014} to establish point correspondence on 3D faces in BU3DFE database. Lu and Jain~\cite{lu2006} presented a multimodal approach for facial feature extraction. Using a face landmark model, the authors detected seven corresponding points on 3D faces using shape index from range images and cornerness from intensity images. Segundo et al.~\cite{segundo2010} combined surface curvature and depth relief curves for landmark detection in 3D faces of the FRGCv2 and BU3DFE databases. They extracted features from the mean and Gaussian curvatures for detecting five landmarks in the nose and eye (high curvature) regions.}  

{Creusot et al.\cite{creusot2013} presented a machine learning approach to detect $14$ corresponding landmarks on 3D faces. They trained multiple LDA classifiers on a set of 200 faces and a landmark model using a myriad of local descriptors. Each landmark detection was treated as a two class classification problem and the final results were fused. This method works well for neutral expression faces of the FRGCv2 and Bosphorus databases. Perakis et al.~\cite{perakis2013} proposed a method to detect landmarks under large pose variations using a statistical Facial Landmark Model (FLM) for the full face and another two FLMs for profile views of the face. Keypoints are detected using  Shape Index and Spin Images and then matched on the basis of minimum combined normalized Procrustes and Spin Image similarity distance from all three FLMs. This method was used to detect eight correspondences in the FRGCv2 and UND Ear databases. Later, the authors proposed a technique~\cite{perakis2014} for fusing features from 2D and 3D data to detect these landmarks with better accuracy than~\cite{perakis2013}.}

{Some methods have also been proposed for generating sparse correspondence for 3D face recognition~\cite{mian2008,smeets2013,berretti2013,li2014}. However, these methods are based on keypoint correspondences that are repeatable only on the same identity.}

\noindent {\textbf{Model based techniques.} These approaches create a morphable model using a sparse set of correspondences and then extend them to dense correspondences.}

{Employing a Point Distribution Model coupled with 3D point signature detection, Nair and Cavallaro~\cite{nair2009} estimated the location of 49 corresponding landmarks on faces. They tested their algorithm on 2,350 faces of the BU3DFE~\cite{yin2006} database and reported a rather high mean landmark localization error.}

\begin{figure}[tb]
\vspace{-0mm}
\centering
\includegraphics[trim = 0pt 0pt 0pt 0pt, clip, width=1\linewidth]{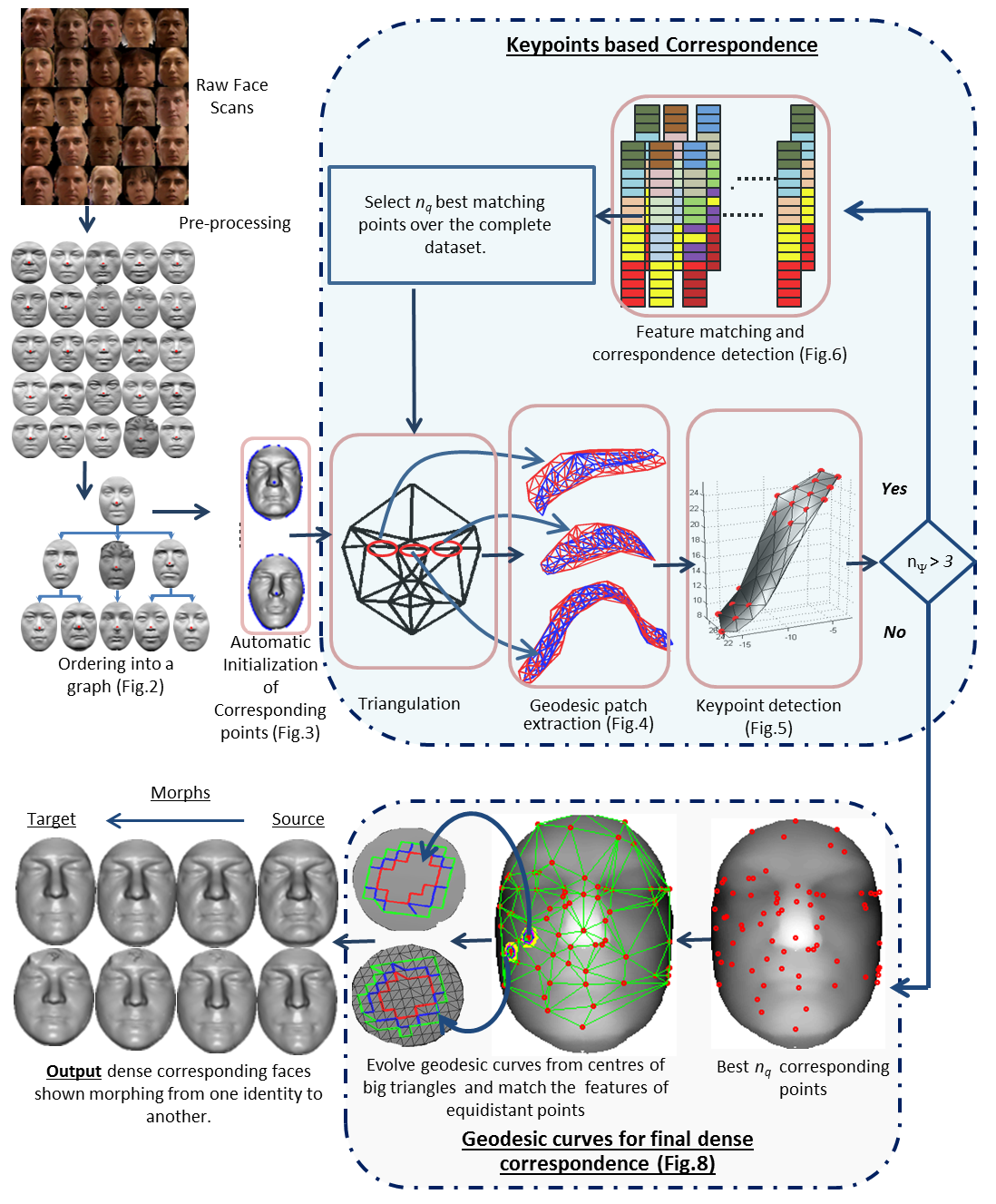}
\vspace{-2mm}
\caption{Block diagram of the presented dense 3D face correspondence algorithm. }
\label{fig:blockdiagram}
\vspace{-7mm}
\end{figure}

{Blanz and Vetter~\cite{blanz1999} proposed a dense correspondence algorithm using optical flow on the texture and the 3D cylindrical coordinates of the face points assuming that the faces are spatially aligned. They constructed a 3D morphable face model from 100 male and female faces each. An arbitrary face was chosen as a reference and the remaining scans were registered to it by iterating between optical flow based correspondence and morphable model fitting. One potential pitfall of the texture based dense correspondence \cite{blanz1999} is that facial texture is not always consistent with the underlying 3D facial morphology e.g. the shape and location of eyebrows. Moreover, this algorithm requires seven manually annotated facial landmarks for initialization. Later, in~\cite{blanz2003,blanz2007} the authors used the 3D morphable model for face recognition. Experiments were performed on only 150 pairs of 3D faces~\cite{blanz2007} from FRGCv2 database, although the total number of scans in the database are $4,007$. The seminal work of Blanz and Vetter~\cite{blanz1999} was extended by Paysan et al.~\cite{paysan2009} in the Basel Face Model (BFM) which used an improved mesh registration algorithm~\cite{amberg2007}. The authors have made their dense correspondence model publicly available which has enabled us to draw comparisons with their model.}

{Passalis et al.~\cite{passalis2005} proposed an Annotated Face Model (AFM) based on an average facial 3D mesh. The model was created by manually annotating a sparse set of anthropometric landmarks~\cite{farkas1994} on 3D face scans and then segmenting it into different annotated areas. Later, Kakadiaris et al.~\cite{kakadiaris2007} proposed elastic registration using this AFM by shifting the manually annotated facial points according to elastic constraints to match the corresponding points of 3D target models in the gallery. Face recognition was performed by comparing the wavelet coefficients of the deformed images obtained from morphing.  Passalis et al.~\cite{passalis2011} further improved the AFM by incorporating facial symmetry to perform pose invariant face recognition. However, the algorithm depends on detection of at least five facial landmarks on a side pose scan.}

{Level set curves were evolved in \cite{gilani2015} to automatically extract seed points and correspondences were established by minimizing the bending energy between patches around seed points of different faces. A morphable model based on the dense corresponding points was then fitted to unseen query faces for transfer of correspondences. The accuracy of landmark localization in this method depends on the number and accuracy of initial seed points.}


\noindent {\textbf{Optimization based techniques.} These methods optimize an objective function to find a mapping between fiducial points. Non-rigid ICP (NICP) is one such technique which formulates deformable registration as an optimization problem consisting of a mesh smoothness term and several data fitting terms~\cite{amberg2007,li2008}. These algorithms require accurate global initialization points ranging from $14$ points~\cite{amberg2007} to $68$ points~\cite{booth2016}. These points are either manually annotated~\cite{blanz1999,amberg2007} or detected automatically using texture~\cite{booth2016}. An extension to this method removes the need for fiducial points but assumes a partial overlap of facial regions~\cite{li2008,li2009}. The alignment between two faces is performed with  a global rigid transformation followed by per-vertex affine transformations that bring the non-rigid shapes into full alignment. Such methods are more suited for time varying deformations of the same identity and often do not result in a bijective (one-to-one) mapping of the vertices. Booth et al.\cite{booth2016} constructed a dense correspondence model of several faces from a propriety dataset by registering the scans to a template mesh using NICP algorithm~\cite{amberg2007} initialized  by $68$ fiducial landmarks detected using texture. Bolkart et al.~\cite{bolkart2016} presented dense correspondence as an optimization problem and used the Minimum Description Length (MDL)~\cite{davies2002} as the objective function. The authors of methods that are based on NICP~\cite{amberg2007,booth2016,paysan2009,li2008,li2009} or other alternative optimization techniques~\cite{bolkart2016} have not reported facial landmark localization results. Hence, it is difficult to perform a direct objective comparison with these methods.}

\vspace{-5mm}
\section{Dense 3D Face Correspondence}

{The overall idea of our system for dense correspondence between 3D face scans, is to begin with a set of {automatically extracted} seed points that represent points matched across all faces in the dataset, and gradually densify the set of matches.  Figure~\ref{fig:blockdiagram} depicts the overall flow of our system.  Here we give an overview of how this proceeds, and then expand the details in the subsections below.}

{We first organize the faces into a tree (section~\ref{sect:Preorg}) based on similarity.  We then seek a set of reliable seed matches (section \ref{sect:convexhull}) from which to begin an iterative densification process.  Each iteration of the densification process (section \ref{sect:TSE}) begins by selecting the current best set of matches (comprising $n_q$ of the full set of $n$ matches)  and forming a triangulation of these points.  Taking each edge of the resulting triangulation in turn, we extract a narrow patch centered on the edge from a pair of faces that are adjacent in the tree.  For each of these patches we find points of distinctive curvature (section \ref{sect:KD}) -- these will be new candidate matches, or \textit{keypoints} -- and compute a 38-dimensional descriptor of the local surface around each keypoint.  Using constrained nearest neighbor we then determine points that match well between the pair of patches (i.e. their descriptors match and they are within a proximity threshold in the patch).  We repeat this process for all parent/child pairs throughout the tree, and eliminate all keypoints that are not consistently matched throughout the tree.  The remaining keypoints that are successfully matched across all faces in the dataset are added to the current set of matches.  At the end of one iteration, when we have cycled through all the triangulated edges, we choose a new best set of $n_q$ matches and repeat the process.}

{Once the search for keypoints is exhausted, further correspondences on facial areas devoid of discriminative points are established by first evolving level set curves and sampling equidistant vertices (See Section~\ref{sect:altmethod}). Feature vectors of these vertices on the reference face are then matched with the remaining faces to establish correspondence as previously stated.} 

{The outcome of this process is a set of densely corresponding 3D faces which we call the Keypoint-based 3D Deformable Model (K3DM).}

\vspace{-5mm}
\subsection{{Preprocessing and Organizing Faces}}
\label{sect:Preorg}
\vspace{-1mm}

{The nose tip of a 3D face is detected automatically following Mian et al.~\cite{mian2007}. Centering a sphere at the nose tip , the face is cropped. The pose of the 3D face is iteratively  corrected to a canonical form using the Hotelling transform~\cite{gonzalez2002}. Next, holes are filled and noise is removed using the gridfit algorithm \cite{john2008}.}

Next, we  pre-organise the face dataset into a graph (in fact, a tree) in which similar faces are ``close'' to one another.  Let $G=(V_g,E_g)$ be a directed graph where each node $V_g$ is a 3D face $\textbf{F}$ from the dataset and each edge $E_g$ connects two nodes $(v_i,v_j)$ of the graph. Each edge of the graph has weight $w$:

\vspace{-3mm}
\begin{equation}
w(v_i,v_j)=\dfrac{\beta_{ij}+\beta_{ji}}{2},
\label{Eq:bendingenergy}
\vspace{-1mm}
\end{equation}
where $\beta_{ij}$ is the amount of bending energy required to deform face $F_i$ to $F_j$ and is measured using the 2D thin-plate spline model~\cite{bookstein1989}. Note that $\beta_{ij} \neq \beta_{ji}$ and $\beta_{ii}=0$. {Since, the faces are already roughly aligned, their nearest neighbor points are taken as approximate correspondences for the purpose of calculating the bending energy.} From $G$, we construct a minimum spanning tree $\Pi=(V_t,E_t)$  using Kruskal's algorithm.  The node with the maximum number of children is taken as the root node.

\begin{figure}[tbp]
\vspace{-0mm}
\centering
\includegraphics[ width=1\linewidth]{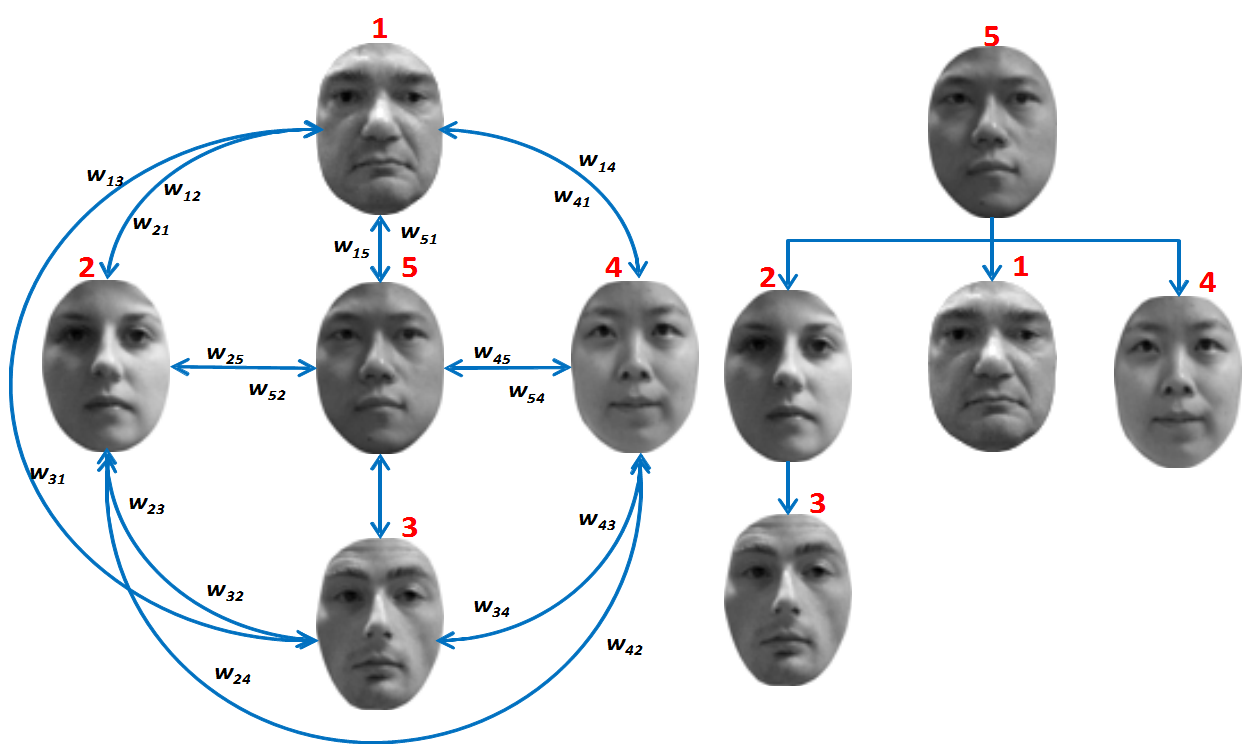}
\vspace{-4mm}
\caption{The directed graph $G=(V_g,E_g)$ (Left) and the {Minimum Spanning Tree} (MST) $\Pi=(V_t,E_t)$ (Right) constructed from five example images of FRGCv2.}
\label{fig:preorg}
\vspace{-4mm}
\end{figure}

The purpose of this pre-organisation is to increase the likelihood of finding point matches between pairs of faces.  A naive approach would be to arbitrarily choose a single (or average) face as reference and find its correspondences to others in the dataset.  But such an approach ignores the proximity between the face instances and the global information underlying the population. {The process and a sample graph are shown in Figure~\ref{fig:preorg}.}

\vspace{-5mm}
\subsection{Sparse Correspondence Initialization}
\label{sect:convexhull}
\begin{figure}[tbp]
\vspace{-0mm}
\centering
\includegraphics[trim = 0pt 10pt 0pt 0pt, clip, width=.95\linewidth]{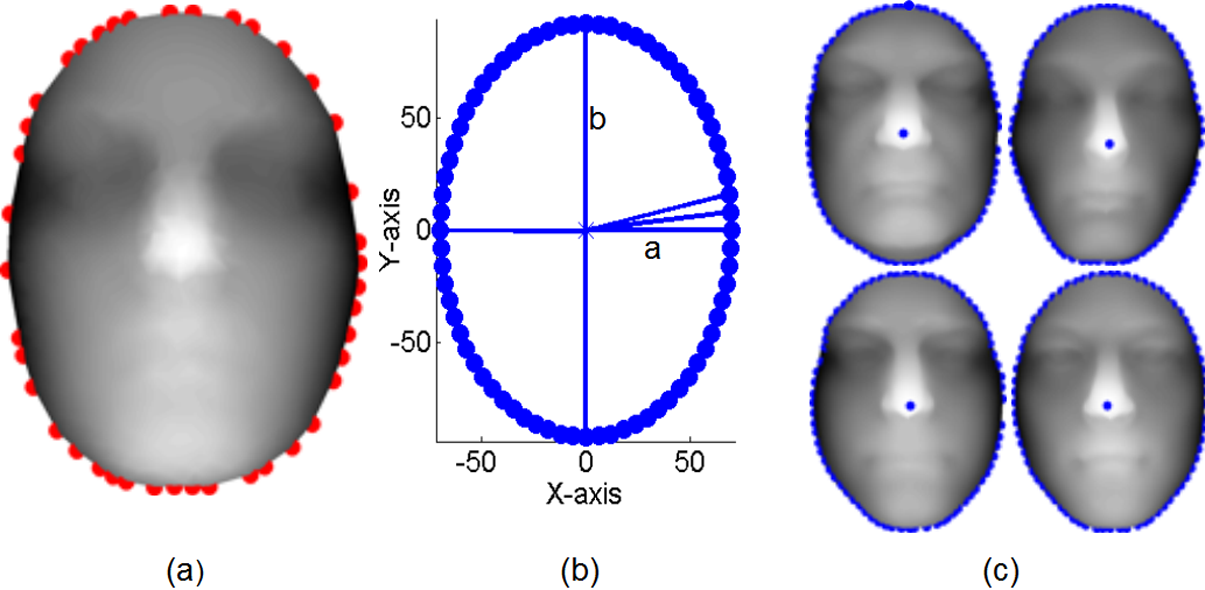}
\vspace{-3mm}
\caption{(a) {Vertices of the 2D-convex hull of the projection} (b) Points sampled at angular intervals of $\pi/36$ (c) Initial sparse correspondence projected on four identities of the FRGCv2 dataset.} 
\label{fig:convhul}
\vspace{-6mm}
\end{figure}

\vspace{-1mm}
{We initialize the correspondences by first automatically establishing a sparse set of seed points.}  We restrict these seed points to those that lie on the roughly ellipse-shaped 2D convex hull of the face i.e. the 2D-hull when the 3D mesh is projected into the $x-y$ plane.  We sample these points at regular angular intervals of $\delta=\pi/36$ {(see Figure~\ref{fig:convhul})}, where the angle $\delta$ is measured at the nose tip.  There is of course no guarantee that in the finite resolution mesh of the face there will be a point at an exact multiple of $\pi/36$, but for each face we choose the nearest point. This yieds a set of 72 3D seed points for each 3D face in the dataset which are used in the first iteration of the triangulation and densification process, as described in the next section. 

\begin{figure}[tbp]
\includegraphics[width=1\linewidth]{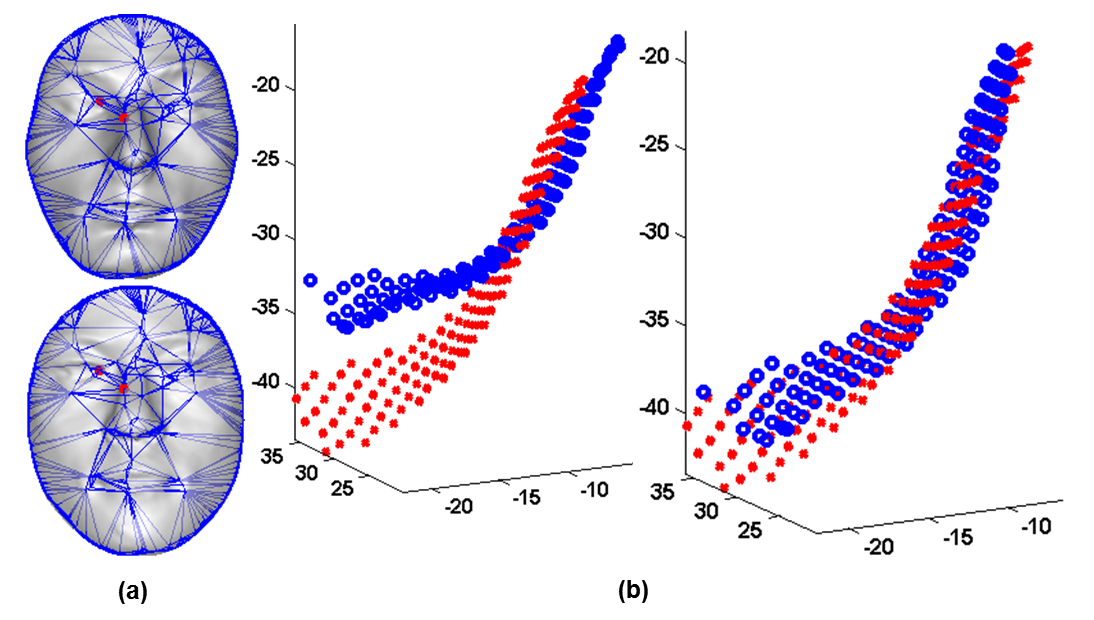}
\caption{Illustration of geodesic patch extraction. (a) Two 3D faces  with triangulation over a few corresponding points from the $2^{nd}$ iteration. Geodesic surface patch is extracted between two sample points shown in red colour. (b) Pointclouds of the geodesic surface patches before and after registration. }
\label{fig:TSE}
\vspace{-5mm}
\end{figure}

\vspace{-5mm}
\subsection{Triangulation and Geodesic Patch Extraction}
\label{sect:TSE}
\vspace{-1mm}

The main part of our algorithm is an iteration that takes the best set of matches that have been established to date, and grows the number of correspondences.  For the first iteration, we use the sparse set of correspondences established as in the previous section, while for subsequent iterations we determine the best set of $n_q$ matches from the full set of $n$ matches as described in section~\ref{sect:FEM}.

In each iteration, given $n_q$ correspondences between $N$ faces, we perform a 2D Delaunay triangulation of the mean $x-y$ locations of the $n_q$ current best matches. This triangulation is then used consistently across all faces.  We then pick a pair of parent/child nodes from the Minimum Spanning Tree $\Pi$, $\mathbf{F}_j$ and $\mathbf{F}_k$.  For both faces in the pair, we extract a narrow surface patch $\mathbf{S}=\{[x_i, y_i, z_i]^T, i=1,\ldots ,m \} \subset \mathbf{F}$, centered on a geodesic curve defined by each triangle edge (see Figure~\ref{fig:TSE}). For the sake of simplicity we call this a geodesic patch.

The (projected) length of the patch is the same as the length of the edge.  The ``narrow'' width is set with reference to the scale of the original face mesh resolution.  More specifically, we set the width to be $5\rho$ where $\rho$ is the average mesh-edge length in the vicinity of the endpont of the edge (note that here the mesh-edges refer to the edges in the original dataset, not the edges of the triangulation used for the densification).  This makes the extraction of the geodesic  patch scale invariant.  The values of $\rho$ for real 3D faces captured with the Minolta\textsuperscript{TM} or the 3dMDface\textsuperscript{TM}  scanners typically range from 1-3mm.  

Finally, we bring the patches $\mathbf{S}_j, \mathbf{S}_k$ into approximate alignment using non-rigid registration~\cite{rueckert1999,dirk2009}.  {The process is shown in Figure~\ref{fig:TSE}.}

\begin{figure}[tbp]
\vspace{-0mm}
\centering
\includegraphics[trim = 0pt 0pt 0pt 0pt, clip, width=0.9\linewidth]{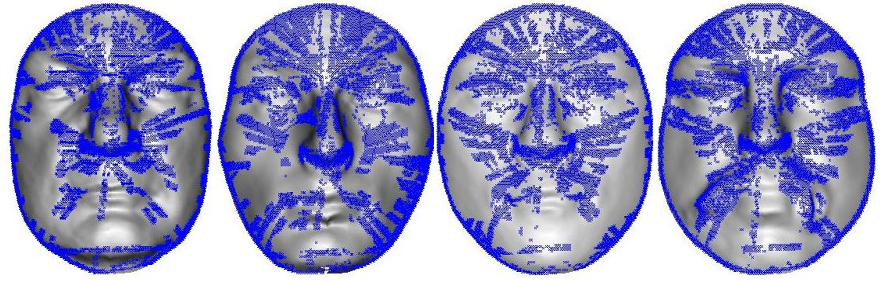}
\vspace{-2mm}
\caption{Illustration of keypoints (not corresponding points) detected along geodesic patches in the tenth iteration of our algorithm. Notice the repeatability of keypoints across the identities. 
}
\label{fig:KD}
\vspace{-7mm}
\end{figure}

\vspace{-3mm}
\subsection{Keypoint Detection on Geodesic Patches}
\label{sect:KD}
\vspace{-0mm}
Our aim now is to establish accurate correspondences between a patch on one face $\mathbf{S}_j$ and its corresponding patch on the other face $\mathbf{S}_k$.  We do this in a fairly standard manner by finding distinctive keypoints, generating a descriptor of the local surface around each point, and establishing matches between points on each patch whose descriptors are sufficiently close.   

More specifically, to find keypoints we consider the surface distinctiveness at each point in the patch.  We do so by calculating the covariance of all the points within a neighborhood of $5\rho$ of the current point, and marking as keypoints any points for which the ratio of the largest two eigenvalues of the covariance exceeds a threshold.  Note that if the neighborhood is uniform these eigenvalues will be equal, and therefore the point is unsuitable as a keypoint.

Figure~\ref{fig:KD} shows keypoints detected by our algorithm in the tenth iteration on four different identities of the FRGCv2 database.

We use the keypoints detected on surface patch $\mathbf{S}_j$ for feature extraction and matching only if an adequate number of keypoints are detected (we use a minimum of three), otherwise, $\mathbf{S}_j$ is not considered to be sufficiently descriptive and the matches are not sought within the patch.

\vspace{-3mm}
\subsection{Feature Extraction and Matching}
\label{sect:FEM}
\vspace{-1mm}

We denote by ${\boldsymbol{\Psi}}_j=[x_i, y_i, z_i]^T, i=1,...,n_\Psi$ the set of keypoints detected on the surface $\mathbf{S}_j$, where $n_\Psi$ is the number of keypoints (likewise for $\boldsymbol{\Psi}_k$). For each keypoint we extract a feature vectors $\mathbf{x}$
which describe the local surface (within $5\rho$) using a set of 3D signature and histogram based descriptors. These descriptors have been widely used in the literature~\cite{creusot2013, guo2013,tombari2010} for automatic object recognition and for landmark detection. We use a combination of many descriptors since the surface patch is quite small and a single descriptor may not capture sufficient information. The list of descriptors is given below:

\begin{figure*}[btp]
	\begin{minipage}[b]{0.325\linewidth}
		\centering
		\includegraphics[trim = 5pt 0pt 20pt 10pt, clip, width=1\linewidth]{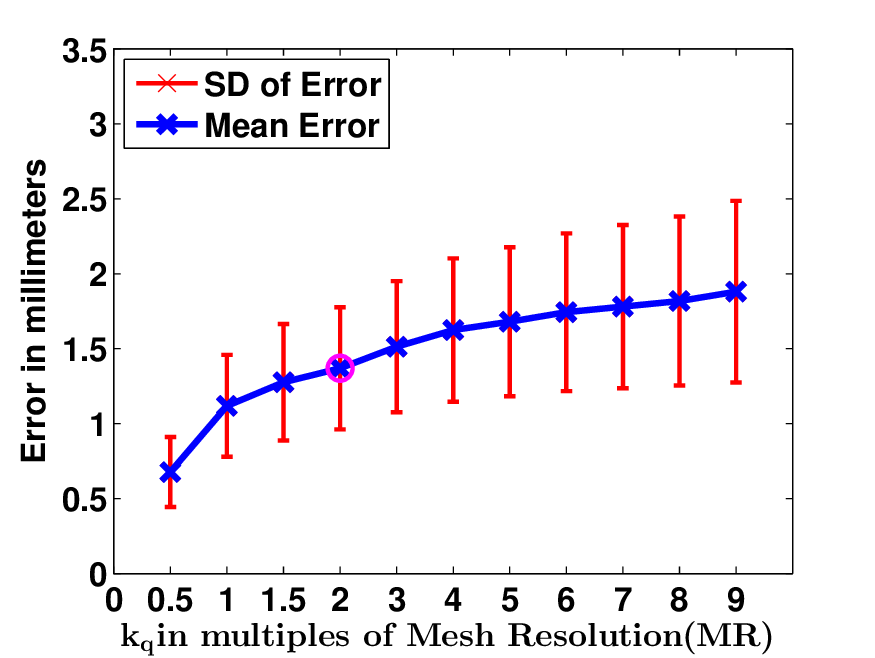}
	\end{minipage}
	\hfill
	\begin{minipage}[b]{.325\linewidth}
		\centering
		\includegraphics[trim = 5pt 0pt 30pt 0pt, clip,width=1\linewidth]{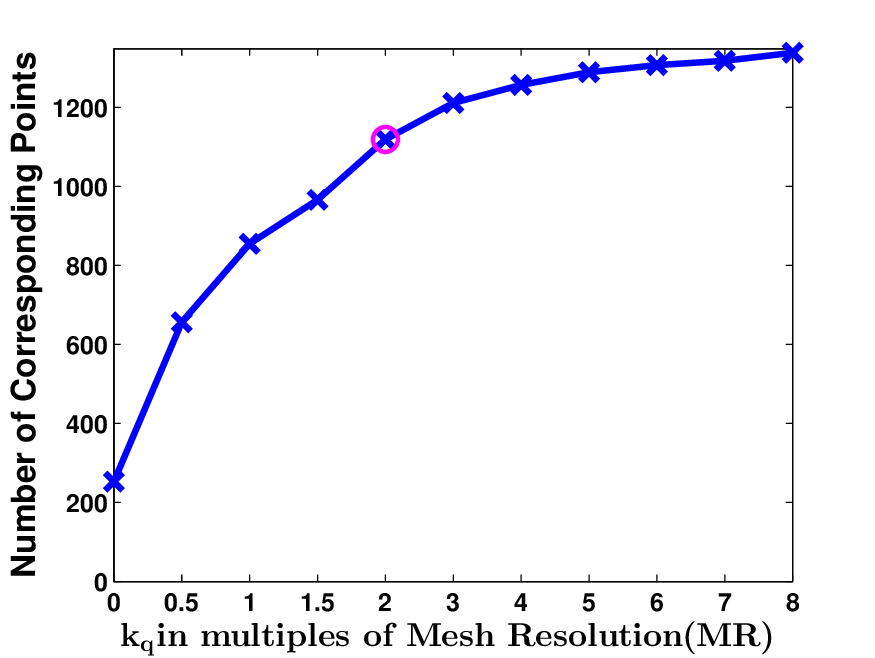}
	\end{minipage}
	\hfill
	\begin{minipage}[b]{0.325\linewidth}
		\centering
		\includegraphics[trim = 10pt 0pt 20pt 0pt, clip, width=1\linewidth]{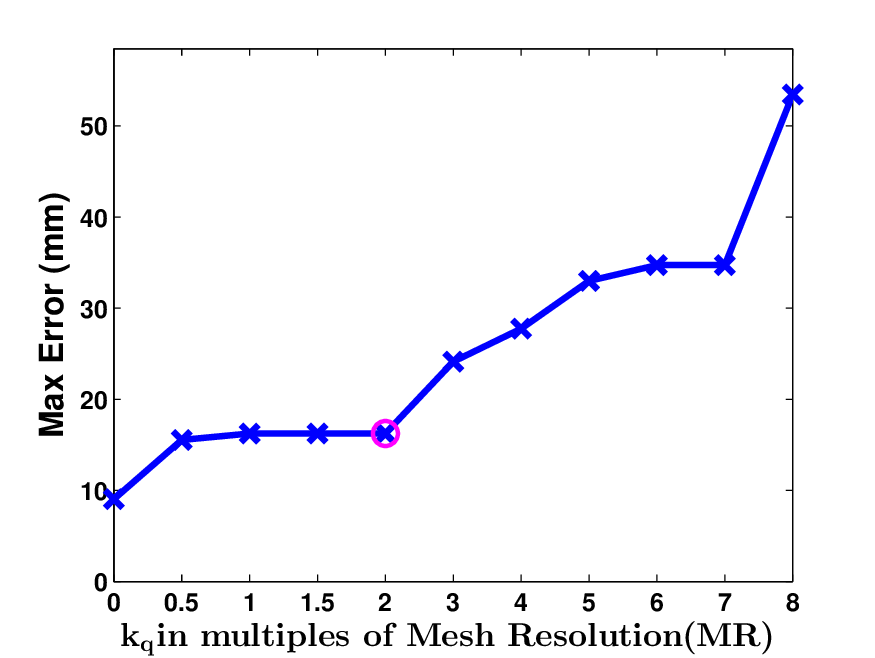}
	\end{minipage}
	\vspace{-2mm}
	\caption{\small{The effect of correspondence quality threshold $k_q$} in the synthetic dataset in the first iteration. (Left) Graph of $k_q$ vs the mean and SD of correspondence localization  error. (Middle) $k_q$ vs the number of correspondences established. (Right) $k_q$ vs the maximum localization  error. For all our experiments we have set $k_q=2\rho$ shown in the graphs in a magenta circle.}
	\label{fig:KqvsErr}
	\vspace{-5mm}
\end{figure*}

\begin{itemize}
\item The spatial location $[x_i, y_i, z_i]^T$.
\item The surface normal $[n_x, n_y, n_z]^T$.
\item The seven invariant moments~\cite{gonzalez2002} of the $3 \times 3$ histograms of the $XY$,$YZ$  and $XZ$ planes. 
\item The central moment $\mu_{mn}$ of order $m+n$ of the histogram matrix $\mathbf{H}$:
\vspace{-1mm}
\begin{equation}
\mu_{mn}=\sum\limits_{i=1}^{\varphi}\sum\limits_{j=1}^{\varphi}(i-\bar{i})^m(j-\bar{j})^n \mathbf{H}(i,j),
\vspace{-1mm}
\end{equation}
\vspace{-0mm}
where $\varphi$ is the total number of points in $\mathbf{H}$, $\bar{i}=\sum\limits_{i=1}^{\varphi}\sum\limits_{j=1}^{\varphi} i \mathbf{H}(i,j)$ and $\bar{j}=\sum\limits_{i=1}^{\varphi}\sum\limits_{j=1}^{\varphi} j \mathbf{H}(i,j)$.

\item 
The mean of the two principle curvatures $\bar{k_1}$ and $\bar{k_2}$ calculated at each point on the extracted local surface
\item The Gaussian Curvature $K=k_1k_2$
\item The Mean Curvature  $H=\dfrac{k_1+k_2}{2}$ 
\item The Shape Index . We use two variants of the shape index which vary from $0$ to $1$ and $-1$ to $1$ respectively,
\\ $s_a=\dfrac{1}{2} - \dfrac{1}{\pi}\arctan \dfrac{k_1+k_2}{k_1-k_2},\quad 0\leq s_a\leq1$  and \\
$s_b=\dfrac{2}{\pi}\arctan \dfrac{k_1+k_2}{k_1-k_2},\quad -1\leq s_b\leq1$. 

\item The Curvedness $c=\sqrt{\dfrac{k_1^2+k_2^2}{2}}$
\item The Log-Curvedness\\ $c_l=\dfrac{2}{\pi}\log\sqrt{\dfrac{k_1^2+k_2^2}{2}}$, 
\item The Willmore Energy $e_w=H^2-K$, 
\item The Shape Curvedness 
$c_s=s_b.c_l$ 
\item The Log Difference Map $m_l=\ln(K-H+1)$.
\end{itemize}

\begin{figure}[tbp]
\vspace{-0mm}
\centering
\includegraphics[trim = 0pt 0pt 0pt 0pt, clip, width=.9\linewidth]{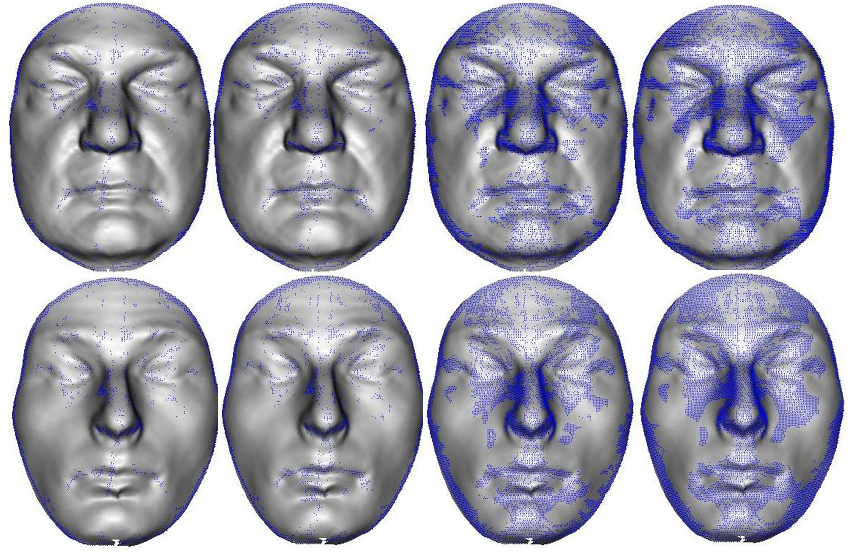}
\vspace{-2mm}
\caption{Correspondence established in $1^{st}$, $4^{th}$, $13^{th}$ and $18^{th}$ iteration  of our algorithm on the first two identities of FRGCv2. Notice how well the points correspond across the identities.}
\label{fig:FEM1}
\vspace{-7mm}
\end{figure}

Using these descriptors, the dimensionality of the final feature vector $\mathbf{x}$ is $38$. These features are extracted over a small enough local surface centered at the keypoint such that they  are repeatable across identities. In contrast, the feature vector extracted by Mian et al.~\cite{mian2008,mian2010} takes the range values of a larger surface (typically 20mm radius) surrounding each keypoint. Hence, their features are repeatable only over the same identity. One of the prerequisites of the techniques that use depth values as features~\cite{mian2008,mian2010,guo2013,tombari2010} is to define a local reference frame for pose invariant matching. {In our case, the features are quasi pose invariant and hence do not require a local reference frame. This is because the pose of each training faces has been iteratively corrected to a canonical form during preprocessing and the features are extracted from a very small patch.}

Next, we perform constrained-NN search between the feature vectors $\mathbf{x}_j$ from $\mathbf{S}_j$ and $\mathbf{x}_k$ from $\mathbf{S}_k$, such that the corresponding points lie within a proximity of $2\rho$ to each other, and their matching score $d(\mathbf{x}_j,  \mathbf{x}_k)$ -- taken to be Euclidean distance between their feature descriptors -- is less than a threshold $k_q$. 
The quality of correspondence varies directly with $k_q$. Higher values of $k_q$ will result in poor corresponding points with large errors, whereas lower values of $k_q$ may reject valid correspondences and hence adversely effect the correspondence density. 
Figure~\ref{fig:KqvsErr} shows the effect of $k_q$ on the correspondence found in our experiments on the synthetic dataset. As we increase the value of $k_q$, the mean localization  error and its standard deviation (SD) increases. Figure~\ref{fig:FEM1} shows the outcome of this step on two identities. 

This process is repeated for all surface patches in a pair of faces, and for all pairs of faces in the MST.  Only points that are matched throughout the MST in the pairwise scheme are retained and these are added to the set of correspondences obtained for the previous iteration.  We then select from the full set of correspondences those which have the smallest matching score $d(\mathbf{x}_j,\mathbf{x}_k)$. We denote the number of selected correspondences by $n_q$ and use a value of $n_q=80$ in our experiments. In order to adequately cover the whole face for the subsequent iteration, we add the original seed points to the $n_q$ points.  Next, we obtain a triangulation of these points on the mean face of the dataset and extract geodesic surface patches as described in Section~\ref{sect:TSE}, repeating the process.

\begin{figure}[t]
\vspace{-0mm}
\centering
\includegraphics[trim = 0pt 8pt 0pt 0pt, clip, width=0.9\linewidth]{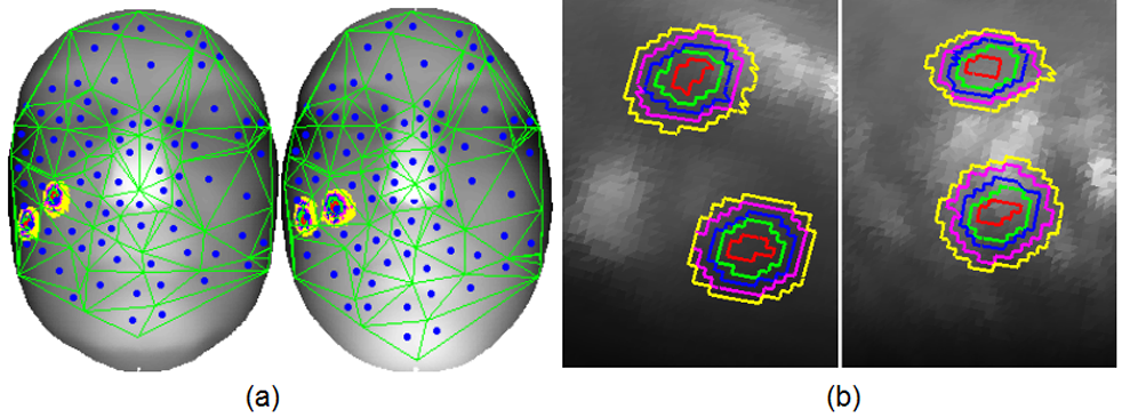}
\vspace{-2mm}
\caption{Correspondence establishment on smooth surfaces. Two faces from an ordered pair with triangulation over $n_q$ best quality corresponding points. Blue dots indicate the centroids of large triangles. Level set based evolution of geodesic curves for the two sample triangles, magnified on the right.}
\label{fig:altmethod}
\vspace{-6mm}
\end{figure}

\vspace{-3mm}
\subsection{Densifying Matches in Uniform Regions}
\label{sect:altmethod}
\vspace{-0mm}

Keypoints, by their very definition concentrate around regions of high curvature/discrimination, such as the mouth, nose, and eyes.  In this section we describe how we establish correspondences in more uniform regions where keypoints cannot be found.
A simple approach to establish dense correspondence in these areas would be to sample them uniformly within triangles of the Delaunay triangulation.  This approach has been used in 2D by Munsell et al.~\cite{munsell2009} who pre-organized the shape instances based on a similarity measure and then established correspondence between pairs of shapes by mapping the points from the source instance to the target instance after minimizing a bending energy.  However, a uniform sampling in the triangle only results in uniform sampling on the face in planar regions.  Instead, we adopt a sampling strategy that respects the underlying surface distances (geodesics) on each face.

After triangulation of the final set of best quality corresponding points, we select large triangles with area greater than $t_a$.  We set $t_a$ to be the mean area of all triangles in the connectivity, an effective and expeditious choice.  From the centroid of each triangle, we evolve a level-set curve, in which the front speed is set to be uniform along a (radial) geodesic.  For convenience we refer to these curves as ``level-set geodesics''. We follow the Fast Marching Method~\cite{sethian2001} and use the implementation given by Peyre~\cite{peyre2011}. We then sample the points along the curve at regular intervals to ensure equidistant points (see Figure~\ref{fig:altmethod}).  Because the evolution speed of the curve is uniform along geodesics, we obtain a uniform sampling on the surface; this is in contrast to uniform sampling within the triangle which would not necessarily be uniform on the surface itself. Although these points are not keypoints, they are repeatable on all 3D faces across identities because they are extracted from triangles whose vertices are corresponding to each other across the dataset. Furthermore, these points are extracted at equal intervals over a small region which is smooth. 

Given this set of points sampled uniformly on the surface, we extract feature vectors and perform pairwise matching as before.  Points whose feature vectors are close enough are retained as matches, with the rest discarded. This is not an iterative process and points are sampled only once from each triangle meeting the threshold criterion. Figure~\ref{fig:altmethod} visually illustrates the process. 

{An alternative method for densifying the matches in the uniform regions could be to register the source and target faces using the NICP algorithm~\cite{amberg2007,li2008} initialized by the correspondences established in the previous section. Once the target face has deformed to the source face, densification of correspondences is achieved by mapping the vertices in uniform regions of the source face to that of the target face. This approach requires tweaking the NICP parameters and the iterative optimization process for non-rigid face deformation tends to be computationally expensive. Our results in Section~\ref{sec:eval-landmark} also show that our feature matching approach achieves higher accuracy and therefore, we use this approach for the remaining part of the paper.}

\vspace{-3mm}
\section{K3DM Fitting and Augmentation}
\label{sect:fitting}
\vspace{-1mm}
The output of the dense correspondence algorithm is the set of  $N$ densely corresponding 3D faces $\widetilde{\mathbf{F}}_j$. {Our objective now is to develop a compact deformable model based on these densely corresponding faces. To do so we take a standard PCA-based approach, and we call the result our Keypoint-based 3D deformable model (K3DM).  More formally, let } $\mathbf{\Upsilon}=[\widetilde{\mathbf{f}}_1,\widetilde{\mathbf{f}}_2,\ldots,\widetilde{\mathbf{f}}_N,]$, where $\widetilde{\mathbf{f}}=[x_1,\ldots,x_p,y_1,\ldots,y_p,z_1,\ldots,z_p]^T$ and $p=1,\ldots,P$. The row mean $\boldsymbol{\mu}_\Upsilon$ of the K3DM is given by,
\vspace{-2mm}
\begin{equation}
\boldsymbol{\mu}_{\Upsilon}=\dfrac{1}{N} \sum\limits_{i=1}^{N} \widetilde{\mathbf{f}}_i
\label{Eq:pdmmean}
\vspace{-2mm}
\end{equation}
\vspace{-0mm}

The row-normalized model $\mathbf{\Upsilon}_m = \Upsilon - \boldsymbol{\mu}_{\Upsilon}$ can be modelled by a multivariate Gaussian distribution and its eigenvalue decomposition is given by,

\vspace{-3mm}
\begin{equation}
\mathbf{USV}^T=\boldsymbol{\Upsilon}_m
\label{Eq:pdmsvd}
\vspace{-2mm}
\end{equation}

\noindent where $\mathbf{US}$ are the principal components (PCs), the columns of $V$ are their corresponding loadings, and $\mathbf{S}$ is a diagonal matrix of eigenvalues. We use only the first $n$ columns of $\textbf{U}$ which correspond to $98\%$ of the energy. 

We propose to deform the statistical model given in~(\ref{Eq:pdmsvd}) into a query face $\mathbf{Q}$ in a two step iterative process, i.e. registration and morphing. Algorithm~\ref{Algo:fitting} gives the details of fitting the deformable model to a query face. Note that we use $\mathbf{Q}$ and $\mathbf{M}$ for the point clouds of the query face and model and use $\mathbf{q}$ and $\mathbf{m}$ for their vectorized versions respectively. The query face after vectorization can be parametrized by the statistical model such that $\mathbf{m}^i=\mathbf{U}\boldsymbol{\alpha}^i+\boldsymbol{\mu}_{\Upsilon}$, where the vector $\boldsymbol{\alpha}^i$ contains the parameters which are used to vary the shape of the model in the $i^{th}$ iteration and $\mathbf{m}^i$ is the vectorized form of the model representing the query face. In the initialization step $\boldsymbol{\alpha}^i$ is set to zero and the deformable model $\mathbf{M}^i$ is characterized  by the mean face of the K3DM. Each iteration begins with a registration step where the input face $\mathbf{Q}$ is registered to the model $\mathbf{M}^i$. This step essentially entails finding an approximate correspondence between the model and the query face and a rigid transformation. Correspondence is established by searching for the Nearest Neighbor (NN) of each point of $\mathbf{M}^i$ in $\mathbf{Q}$ using the k-d tree data structure~\cite{bentley1975}. 
Let $\mathbf{d}$ represent the NN Euclidean distance between the corresponded query face and the model such that $d_j=\left \|\widetilde{\mathbf{Q}_j}^i-\mathbf{M}_j^i\right\|_2$. We define outliers as points on $\widetilde{\mathbf{Q}}$ whose NN distance with $\mathbf{M}^i$ is greater than a threshold $t_c$ where $t_c=\overline{\mathbf{d}}+3\sigma_d$ and exclude them from registration. This step ensures that the outliers do not affect the registration process. Next, the query face is translated to the mean of the model and is rotated to align with $\mathbf{M}^i$. We denote the corresponded and registered query face by $\mathbf{Q}_r$.

\begin{figure}[t]
\vspace{-0mm}
\centering
\includegraphics[trim = 0pt 5pt 0pt 0pt, clip, width=1\linewidth]{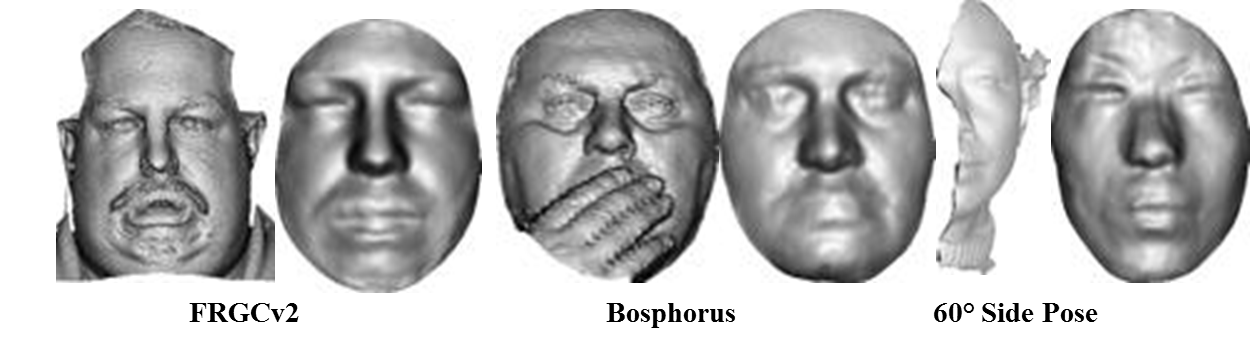}
\vspace{-6mm}
\caption{K3DM fitting results on three datasets. The first scan for each dataset is the raw input while the second scan is the fitted model. The $60^{\circ}$ side pose scan has been rotated to highlight the partial data.}
\label{fig:fitting}
\vspace{-6mm}
\end{figure}

In the next step, the model $\mathbf{M}^i$ is deformed to fit the registered query face $\mathbf{Q}_r$ such that, 
\vspace{-2mm}
\begin{equation}
\hat{\boldsymbol\alpha} ^{i} \Leftarrow \min _{\boldsymbol\alpha ^{i}}\left\| \mathbf{U}^\ast \boldsymbol\alpha^i +\boldsymbol\mu _{\Upsilon }-\mathbf{q}_r\right\|_2+\lambda\|\boldsymbol\alpha^i-\boldsymbol\alpha^{i-1}\|_2
\label{Eq:pdmmodel}
\vspace{-2mm}
\end{equation}
\vspace{-0mm}
and $\mathbf{m}^i=\mathbf{U}\hat{\boldsymbol\alpha} ^{i}+\boldsymbol{\mu}_{\Upsilon}$. The $^\ast$  denotes that only those points (rows of $\mathbf{U}$ and $\boldsymbol{\mu}_{\Upsilon}$) are considered which satisfy the threshold $t_c$. The second term in~(\ref{Eq:pdmmodel}) puts a constraint on deforming the model. The applied condition is intuitive because we want to partially deform the model in each iteration such that the model approximates the query face in small steps. 
The iterative procedure is terminated when the residual error $\|\mathbf{m}^i-\mathbf{q}_r\|_2 \le \epsilon_f$. In all of our experiments $\lambda$ was set to $0.8$ and $\epsilon_f=10^{-4}$.

\floatname{algorithm}{Algorithm}
\renewcommand{\algorithmicensure}{\textbf{Initialization:}}
\begin{algorithm} 
\caption{K3DM Fitting} 
\label{Algo:fitting} 
\begin{algorithmic}[1]
\REQUIRE  $\boldsymbol{\Upsilon_m}=[\widetilde{\mathbf{f}}_1,\widetilde{\mathbf{f}}_2,\ldots,\widetilde{\mathbf{f}}_N]-\boldsymbol{\mu}_{\Upsilon}$ and Query Face $\mathbf{Q}=[x_p,y_p,z_p]^T$ where $p=1,\ldots,P_q$.
\ENSURE 
\STATE Iteration: $i = 0$ and $\epsilon^0=1$
\STATE $\mathbf{USV}^T=\boldsymbol{\Upsilon}_m$
\STATE $\boldsymbol{\alpha}^i=0$ and  $\mathbf{m}^i=\mathbf{U}\boldsymbol{\alpha}^i+\boldsymbol{\mu}_{\Upsilon}$
\WHILE {$\epsilon^i>\epsilon_f$}
\STATE Update iteration: $i = i +1$\\
\STATE $\widetilde{\mathbf{Q}}=\mathbf{Q}{\Leftarrow }\mathbf{M}^i$ (NN using k-d tree)
\STATE $\widetilde{\mathbf{Q}}'=\{\widetilde{\mathbf{Q}}\left \|\widetilde{\mathbf{Q}}^i-\mathbf{M}^i\right\|_2<\overline{\mathbf{d}}+3\sigma_d\}$
\STATE $\mathbf{Q}_r=\widetilde{\mathbf{Q}}' \mathbf{R}+\mathbf{t}$ (Registration step)
\STATE $\mathbf{U}^\ast {\Leftarrow}$ \{ $\mathbf{U}\|$ rows of $\mathbf{U}$ correspond to $\widetilde{\mathbf{Q}}'$ \}
\STATE $\hat{\boldsymbol\alpha} ^{i} {\Leftarrow} \min _{\boldsymbol\alpha _{i}}\left\| \mathbf{U}^\ast \boldsymbol\alpha^i +\boldsymbol\mu _{\Upsilon }-\mathbf{q}_r\right\|_2+\lambda\|\boldsymbol\alpha^i-\boldsymbol\alpha^{i-1}\|_2$\\
\STATE $\mathbf{m}^i=\mathbf{U}\hat{\boldsymbol\alpha} ^{i}+\boldsymbol{\mu}_{\Upsilon}$
\STATE $\epsilon^i=\|\mathbf{m}^i-\mathbf{q}_r\|_2$
\ENDWHILE
\RETURN $\mathbf{Q}_r, \hat{\boldsymbol\alpha}, \mathbf{m}$
\end{algorithmic} 
\end{algorithm} 
\vspace{-1mm}
\floatname{algorithm}{Algorithm}
\renewcommand{\algorithmicensure}{\textbf{Initialization:}}
\begin{algorithm} 
\caption{K3DM Augmentation} 
\label{Algo:DM_Augment} 
\begin{algorithmic}[1]
\REQUIRE  
$\boldsymbol{\Upsilon}=[\widetilde{\mathbf{f}}_1,\widetilde{\mathbf{f}}_2,\ldots,\widetilde{\mathbf{f}}_N]$ and a batch of input 3D faces $F_M=\{\mathbf{f}_1,\mathbf{f}_2,\ldots,\mathbf{f}_M\}$, $M\ge1$. 
\ENSURE
\STATE Pre-organize the $M$ faces in a Minimum Spanning Tree $\Pi=(V_t,E_t)$
\FOR {each 3D face $\mathbf{f}_i$ in $\Pi$}
	\STATE $\widetilde{\mathbf{f}}_i=\textrm{fit\_K3DM}(\boldsymbol{\Upsilon},\mathbf{f}_i)$ 
	\STATE $\boldsymbol{\Upsilon}=[\widetilde{\mathbf{f}}_1,\widetilde{\mathbf{f}}_2,\ldots,\widetilde{\mathbf{f}}_N,\widetilde{\mathbf{f}}_i]$ 
	\STATE Increment number of faces in the model
	\STATE $\boldsymbol{\mu}_{\Upsilon}=\dfrac{1}{N} \sum\limits_{n=1}^{N} \mathbf{\Upsilon}_n$
\ENDFOR
\RETURN$\boldsymbol{\Upsilon}=[\widetilde{\mathbf{f}}_1,\widetilde{\mathbf{f}}_2,\ldots, \widetilde{\mathbf{f}}_{N+M}]$
\end{algorithmic} 
\end{algorithm} 
From a practical perspective, there is usually a need to augment an existing dense correspondence model with new 3D faces. In the following, we present a K3DM augmentation algorithm to achieve this objective.
Given the K3DM and a batch of \textit{M} new 3D faces, we compute the bending energy required to deform the mean face of the K3DM to each of the new faces. This information is employed to  organize the \textit{M} faces in a Minimum Spanning Tree as outlined in Section~\ref{sect:Preorg}. Traversing from the root node (mean face), the K3DM is morphed  into each child node using the model fitting procedure given in Algorithm~\ref{Algo:fitting}. The resulting corresponded 3D face of the input identity is added to the K3DM. 
Algorithm~\ref{Algo:DM_Augment} gives the details of our model augmentation technique.


\begin{figure}[t]
\vspace{-0mm}
\centering
\includegraphics[trim = 0pt 0pt 0pt 0pt, clip, width=1\linewidth]{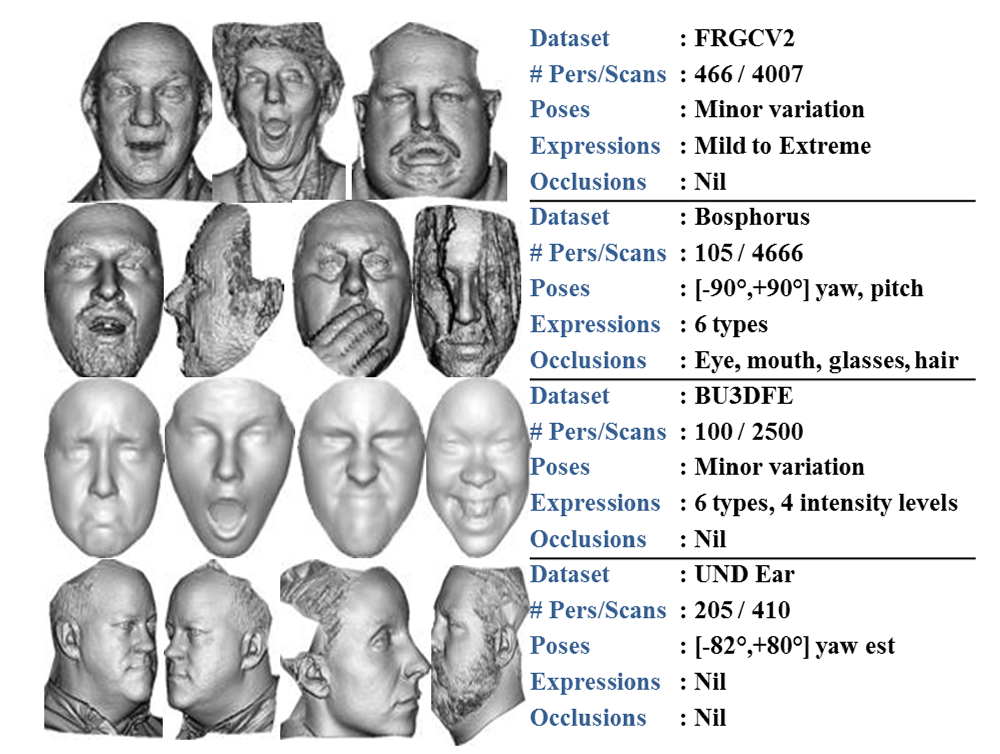}
\vspace{-6mm}
\caption{Sample images and details of our four experimental datasets.}
\label{fig:datasets}
\vspace{-6mm}
\end{figure}

\vspace{-4mm}
\section{Experimental Setup}
\label{sect:exp}
\vspace{-1mm}
We have carried out extensive experiments on synthetic and real data. 
Below are the details of the datasets used, evaluation criteria and the experiments performed.

\vspace{-4mm}
\subsection{Datasets Used}
\vspace{-0mm}

Our synthetic dataset consists of  100 3D faces generated from the Facegen software\footnote{Singular Inversions, ``Facegen Modeller'', www.facegen.com}. 
Facegen has been used by scientists in the field of neuroscience and social cognition to generate synthetic faces for replicating human stimuli~\cite{todorov2008,oosterhof2008}. 
The 100 faces are in perfect correspondence with each other and hence provide the ground truth. Each face has 3,727 vertices and 7,179 triangles. 
{For experiments on real 3D faces, we used the FRGCv2~\cite{phillips2005}, Bosphorus~\cite{savran2008}, BU3DFE~\cite{yin2006} and side pose scans of the UND Ear database Collections F~\cite{yan2005} and G~\cite{yan2006}. Some sample images and details of these datasets are given in Figure~\ref{fig:datasets}. The purpose of using such diverse datasets was to evaluate the performance of our proposed technique for partial data, occlusion, expression and pose invariance.}

\vspace{-4mm}
\subsection{Evaluation Criteria}
\begin{figure}[t]
\vspace{-0mm}
\centering
\includegraphics[trim = 0pt 0pt 0pt 0pt, clip, width=0.9\linewidth]{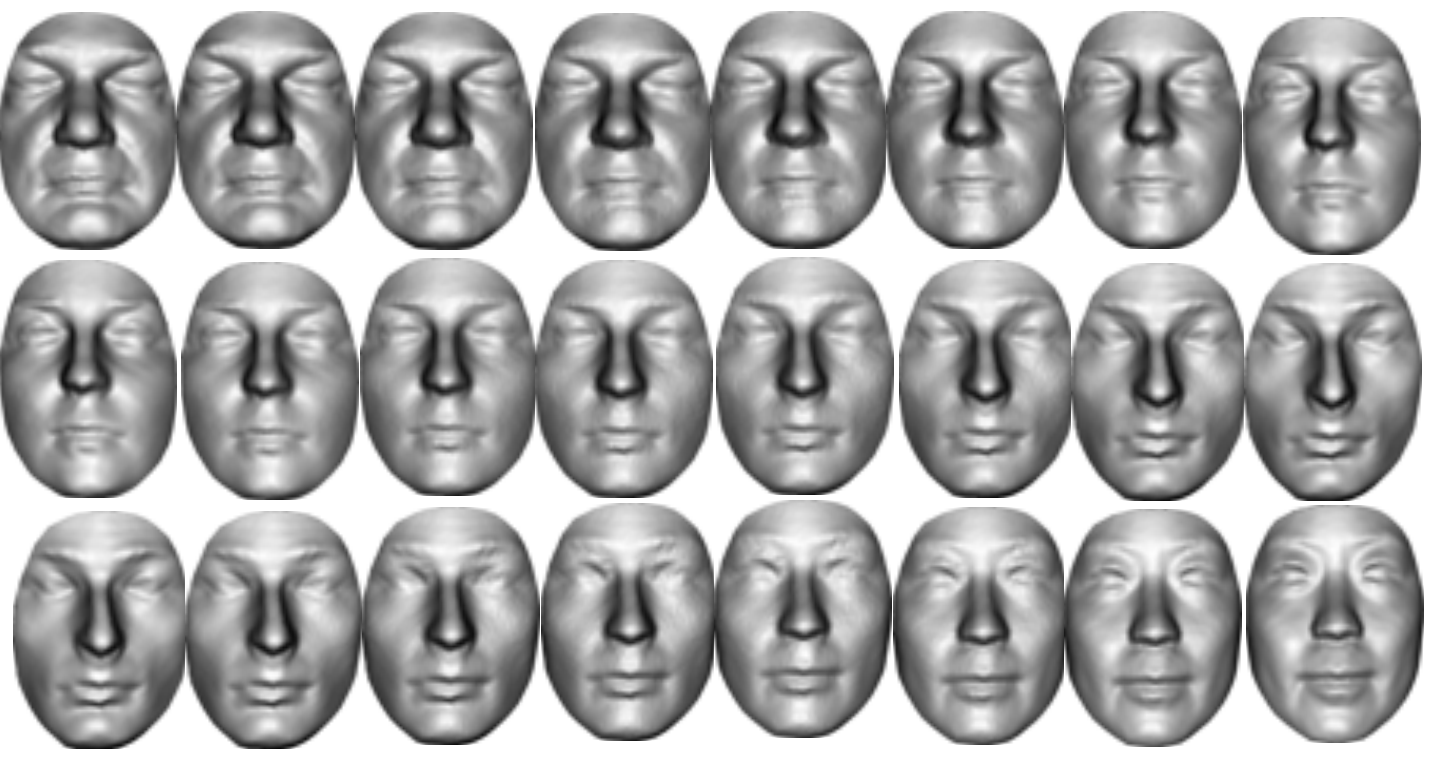}
\vspace{-3mm}
\caption{Qualitative results of our dense correspondence algorithm on the first 3 identities of FRGCv2. The first face in each row is the source and the last face is the target.} 
\label{fig:morph}
\vspace{-7mm}
\end{figure}

\vspace{-1mm}
Figure~\ref{fig:morph} shows qualitative results of our dense correspondence algorithm. The smooth transition between different faces is indicative of accurate correspondences~\cite{zhang2008,kraevoy2004}. We have included a video of morphings in the supplementary material. 

Objective evaluation of dense correspondence algorithms on real data is difficult due to the unavailability of the ground-truth shape correspondences~\cite{munsell2008}. One solution is  to use synthetic data where correspondences are known a priori. We used the synthetic 3D face dataset as ground truth for our evaluations. To the best of our knowledge, this is the first time synthetic 3D face images have been used to evaluate results of a dense correspondence algorithm in terms of mean localization  error of the correspondences.  {This dataset and protocol was also used to evaluate the efficacy of individual modules of our algorithm.}

In the case of real data, the accuracy of the dense correspondence can be measured together with the deformable model fitting algorithm by measuring the accuracy of landmark localization and face recognition. Results are expected to be better when the underlying models have accurate dense correspondences. Hence, we used our dense correspondence models and fitting algorithm in these applications and evaluated the results. In all tables, we have highlighted the best and the second best result in that category. Note that the main focus of this paper is to propose a dense 3D face correspondence algorithm. Experiments on landmark localization  and face recognition have been carried out to validate the accuracy of the correspondences.

We create separate dense correspondence models from the FRGCv2, Bosphorus and BU3DFE datasets and denote them with K3DM$_{FR}$, K3DM$_{BO}$ and K3DM$_{BU}$ respectively.  We compare our results to the Basel Face Model (BFM) proposed by Paysan et al.~\cite{paysan2009}. {We also establish dense correspondence using our NICP variant for densifying the initial keypoints based feature matches (see Section~\ref{sect:altmethod}). The algorithm is initialized by the correspondences found in Section~\ref{sect:FEM} and the variant is referred as K3DM-NICP.}


\begin{table}[b]
\vspace{-3mm}
\centering
       \scriptsize
       \caption{Module wise mean and SD of localization error (mm) on 2,246 vertices of the Synthetic dataset.}
        \setlength{\tabcolsep}{3.5pt}
        	 \def\pz{\phantom{0}}
        	 \renewcommand{\arraystretch}{1.1}
        \vspace{-1mm}
    \begin{tabular}{|l|c|}
    \hline
    {\textbf{Excluded Module(s)}} & {\textbf{mean $\pm$ std}} \\
    \hline
    Organising faces into a graph & 2.16 $\pm$ 2.8 \\
    Keypoint detection& 3.06 $\pm$ 5.1 \\
    Feature matching  & 3.61 $\pm$ 6.8 \\
    Keypoint detection and feature matching & 4.78 $\pm$ 7.3 \\
    Selecting best matches in each iteration & 2.61 $\pm$ 3.4 \\
    No modules excluded & \textbf{1.28 $\pm$ 2.2} \\
    \hline
    \end{tabular}%
  \label{tab:Module_LLE}%
  \vspace{-0mm}
\end{table}%

\vspace{-5mm}
\section{Results and Analysis}
\vspace{-0mm}
\subsection{Landmark Localization}
\label{sec:eval-landmark}
\vspace{-1mm}

\begin{figure*}[t]
\vspace{-0mm}
\begin{minipage}[b]{0.24\linewidth}
  \centering
\includegraphics[trim = 0pt 0pt 30pt 0pt, clip, width=1\linewidth]{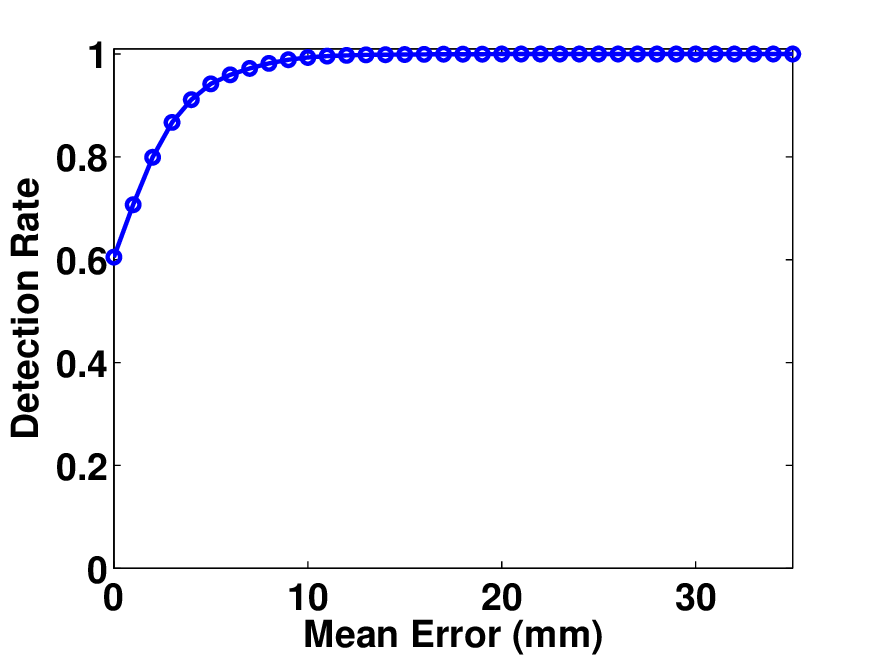}
  \centerline{\small{(a) Synthetic Dataset }}\medskip
\end{minipage}
\hfill
\begin{minipage}[b]{0.245\linewidth}
  \centering
\includegraphics[trim = 5pt 5pt 20pt 0pt, clip, width=1\linewidth]{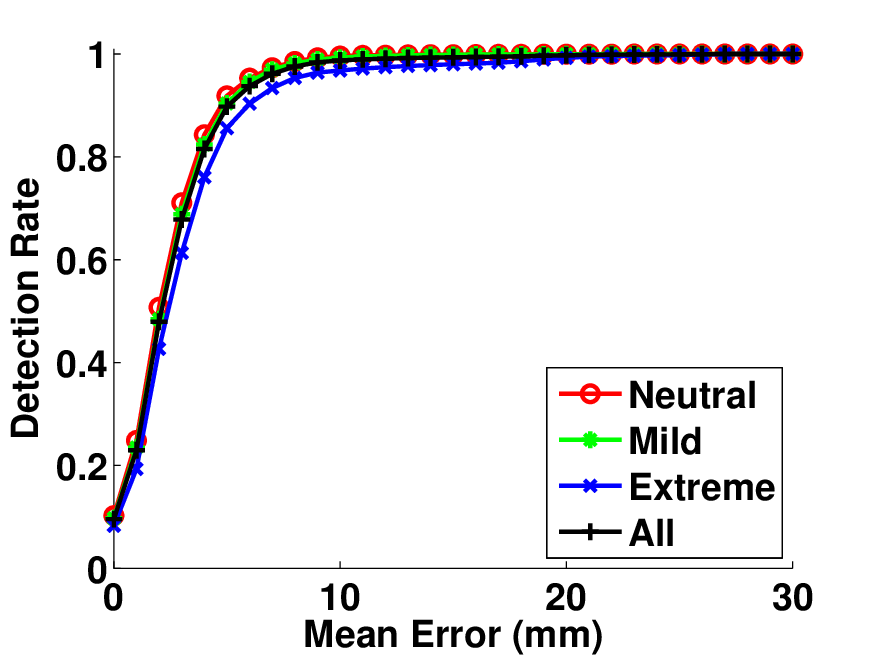}
  \centerline{\small{(b) FRGCv2 }}\medskip
\end{minipage}
\hfill
\begin{minipage}[b]{.245\linewidth}
  \centering
\includegraphics[trim = 5pt 5pt 30pt 0pt, clip,width=1\linewidth]{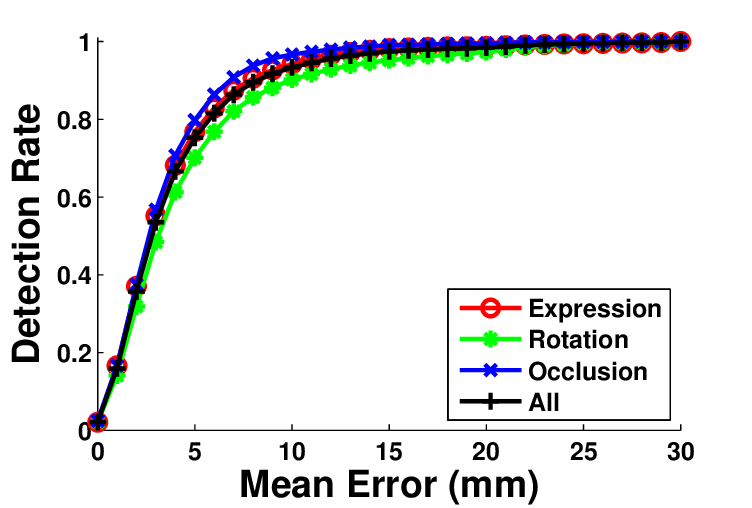}
  \centerline{\small{(c) Bosphorus }}\medskip
\end{minipage}
\hfill
\begin{minipage}[b]{.24\linewidth}
  \centering
\includegraphics[trim = 5pt 5pt 30pt 0pt, clip,width=1\linewidth]{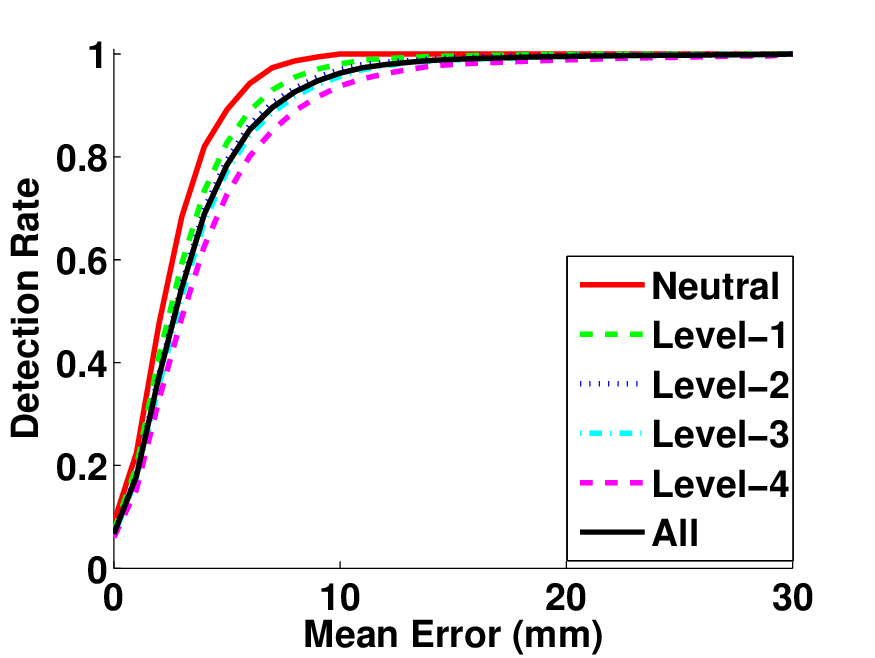}
  \centerline{\small{(d) BU3DFE }}\medskip
\end{minipage}
\hfill
\vspace{-4mm}
\caption{\small{Results of dense correspondence: (a-d) Cumulative localization  error distribution plots on the Synthetic (2,246 vertices), FRGCv2 (14 landmarks), Bosphorus (14 landmarks) and BU3DFE (12 landmarks) datasets. }}
\label{fig:LLE_All}
\vspace{-5mm}
\end{figure*}

\textbf{Synthetic Dataset}: First, we present the evaluation of our algorithm on synthetic data. We establish dense correspondence on 100 synthetic 3D faces using our proposed algorithm and report the mean and the standard deviation (SD) of the localization  error with respect to the ground truth.  
The original synthetic dataset contains 3,727 vertices for each 3D face. Our proposed method is successful in establishing dense correspondence over 2,246 vertices ($60\%$ of the original) with a mean localization  error of $1.28mm$ and SD $\pm 2.2mm$. Correspondence within 10mm is established on $99.33 \%$ vertices. Figure~\ref{fig:LLE_All}(a) shows a plot of the cumulative distribution of correspondences within a given error distance. {We also establish dense correspondence over 2,341 vertices of the synthetic dataset using our K3DM-NICP variant (see Section~\ref{sect:altmethod}). That method results in a mean localization error of $1.30mm$ with $\pm2.3mm$ standard deviation.} 

To ascertain the contribution of different components, we repeat our experiments by removing different components (see Figure.~\ref{fig:blockdiagram}) from our algorithm. The results in Table~\ref{tab:Module_LLE} show that the combination of all components/modules  gives the best results.

\begin{table*}[htbp]
\vspace{-5mm}
  \centering
  \scriptsize
   \caption{Comparative results of the mean and SD (mm) of landmark localisation error on FRGCv2 dataset. A \textquoteleft-\textquoteright \ denotes that the authors have not detected this particular landmark. Ex/En-outer/inner eye corner, N-nosebridge saddle, Prn-nosetip, Ac-nose curvature, Ch-mouth corner, Ls/Li upper/lower lip midpoint, Pg-chintip, Sn-nasal base.}
    \vspace{-3mm}
    \setlength{\tabcolsep}{0.5pt}
    \def\pz{\phantom{0}}
    \renewcommand{\arraystretch}{1.2}
    \begin{tabular}{|l|c|c|c|c|c|c|c|c|c|c|c|c|c|c|}
    \hline
    {\textbf{Author}} & \textbf{Images} &\textbf{Ex(L)} & \textbf{Ex( R )}  & {\textbf{En(L)}} & {\textbf{En( R)}} & {\textbf{N}} & {\textbf{Prn}} & \textbf{Ac$\mbox{*}$}  & {\textbf{Ch$\mbox{*}$}}  &{\textbf{Ls}}  & {\textbf{Li}} & {\textbf{Pg}} & {\textbf{Sn}}  & {\textbf{Mean}}   \\ 
    \hline
    Lu~\cite{lu2006}    & \pz{}676   & 9.5  $\pm$ 17.1 & 10.3 $\pm$ 18.1 & 8.2  $\pm$17.2 & 8.3  $\pm$ 17.2 & -  & 8.3  $\pm$ 19.4 &  -  & 6.1 $\pm$ 17.4   & -         & -         & -         & -          & 8.1 $\pm$ 17.7 \\
    
    Segundo~\cite{segundo2010} & 4007  & -         & -         & 3.7  $\pm$ 2.3  & 3.4  $\pm$ 2.3  & -          & 2.8  $\pm$ 1.4  & 5.3  $\pm$ 1.9     & -  & -     &              & -     & -     & 4.1 $\pm$ 1.9 \\
    
    Perakis~\cite{perakis2013}&  \pz{}975   & 5.6  $\pm$ 3.1  & 5.8  $\pm$ 3.4  & 4.2   $\pm$ 2.2  & 4.4   $\pm$ 2.5  & -     & -     & 4.1   $\pm$ 2.2  &       5.5   $\pm$ 2.4   & -     & -        & 4.9   $\pm$ 3.7     & - & 5.0   $\pm$ 2.7 \\
    
    Cruesot~\cite{creusot2013}  & 4007  & 5.9  $\pm$ 3.1  & 6.0  $\pm$ 3.0  & 4.3  $\pm$ 2.4  & 4.3   $\pm$ 2.0  & 4.2   $\pm$ 2.0  & 3.4   $\pm$ 2.0  & 4.8  $\pm$ 3.6  & 5.5   $\pm$ 3.5  & 4.2  $\pm$ 3.2  & 5.5   $\pm$ 3.9 & 7.3   $\pm$ 7.4  & 3.7   $\pm$ 3.1  & 5.0  $\pm$ 3.3 \\
    
    Perakis~\cite{perakis2014} &  \pz{}975   & \multicolumn{1}{l|}{~4.7} & \multicolumn{1}{l|}{~5.4}  & \multicolumn{1}{l|}{~4.0} & \multicolumn{1}{l|}{~4.1} & \multicolumn{1}{l|}{~3.7}&    \multicolumn{1}{l|}{~4.3}   & -     & \multicolumn{1}{l|}{~4.1}  & -     & -    &  &        & \multicolumn{1}{l|}{~4.3}    \\
   
    Sukno~\cite{sukno2015}  & 4007  & 4.7  $\pm$ 2.7  & 4.6  $\pm$ 2.7      & 3.5   $\pm$ 1.7  & 3.6   $\pm$ 1.7     & \textbf{2.5   $\pm$ 1.6}  & 2.3  $\pm$ 1.7 & \textbf{2.6}  $\pm$ {1.4}  & 3.9   $\pm$ 2.8      & 3.3   $\pm$ 1.8 & 4.6   $\pm$ 3.4  & 4.9   $\pm$ 3.5  & \textbf{2.7  $\pm$ 1.1}  & 3.5   $\pm$ 2.4 \\
    
    Gilani~\cite{gilani2015} & 4007  & 4.5  $\pm$ 2.9  & 3.7 $\pm$ 2.8  & 3.1   $\pm$ 2.1  & 2.7   $\pm$ 2.1  & 3.6   $\pm$ 2.0  & 2.7   $\pm$ 2.5  & 4.2   $\pm$ 3.2  & 4.8   $\pm$ 2.1  & 3.3   $\pm$ 3.7  & 4.0   $\pm$ 3.8  & 4.2   $\pm$ 3.3  & 4.1   $\pm$ 3.1  & 3.9   $\pm$ 2.8 \\
    
    BFM~\cite{paysan2009}   & 4007  & \textbf2.2 $\pm$ 2.5  & 2.7  $\pm$ 1.8  & 2.5   $\pm$ 2.1  & 2.9   $\pm$ 2.2  & 3.2   $\pm$ 2.2  & 2.3  $\pm$ 2.0  & 8.3  $\pm$ 2.9  & \textbf{2.6   $\pm$ 2.9}  & {2.6  $\pm$ 2.2}  & \textbf{3.8  $\pm$ 3.7}  & 4.2   $\pm$ 3.8  & 3.8   $\pm$ 3.6  & 3.7   $\pm$ 2.7 \\
    
   	{K3DM-NICP} & {4007} & {2.8 $\pm$ 2.2} & {2.5 $\pm$ 1.8} & {2.7 $\pm$ 1.8} & {2.6 $\pm$ 1.1} & {2.6 $\pm$ 1.7}  & {2.4 $\pm$ 1.9}  & {3.3 $\pm$ 2.5}  & {2.7 $\pm$ 1.8}  & {2.6 $\pm$ 3.2}  & {4.2 $\pm$ 3.4} & {4.2 $\pm$ 3.3} & {3.5 $\pm$ 1.4} & {3.3 $\pm$ 2.3} \\ 
    
    \textbf{K3DM$_{FR}$} & 4007  & \textbf{2.6  $\pm$ 2.1} & \textbf{2.4 $\pm$ 1.7} & \textbf{2.4  $\pm$ 1.6} & \textbf{2.4 $\pm$ 0.9} & \textbf{2.5}  $\pm$ {1.5} & \textbf{2.2   $\pm$ 1.8}  & \textbf{3.0   $\pm$ 2.4}  & \textbf{2.5  $\pm$ 1.8}  & \textbf{2.4   $\pm$ 3.1}  & 4.1 $\pm$ 3.3 & \textbf{4.1   $\pm$ 3.3} & \textbf{3.4   $\pm$ 1.1} & \textbf{2.9 $\pm$ 2.1} \\
    
    K3DM$_{BU}$ & 4007  & 2.7  $\pm$ 2.4  & \textbf{2.3  $\pm$ 1.9}  & \textbf{2.4   $\pm$ 1.9}   & \textbf{2.5   $\pm$ 1.8} & 2.8  $\pm$ 1.8  & 2.6   $\pm$ 1.8 & 6.1   $\pm$ 2.7  & 4.2  $\pm$ 3.1   &{2.9   $\pm$ 3.3}  & 4.6   $\pm$ 3.9 & \textbf{4.1  $\pm$ 3.4}  & 3.6   $\pm$ 2.9  & 3.6  $\pm$ 2.6 \\
    
    K3DM$_{BO}$ & 4007  & 2.6  $\pm$ 2.2  & 2.4  $\pm$ 1.9  & 2.8   $\pm$ 2.0  & 2.9   $\pm$ 2.0  & 3.2   $\pm$ 2.2  & \textbf{2.3  $\pm$ 2.1}  & 8.3  $\pm$ 3.4  & 3.1   $\pm$ 2.7   & \textbf{2.5  $\pm$ 2.4}  & \textbf{3.5  $\pm$ 3.7}  & 4.1   $\pm$ 3.9  & 3.8   $\pm$ 3.6  & 3.8   $\pm$ 2.7 \\
    \hline 
    \multicolumn{15}{l}{$\mbox{*}$ Results have been averaged for left and right corners of nose and mouth. }\\
    \end{tabular}%
    \vspace{-4mm}
  \label{tab:FRGC_LLE_CompC}%
\end{table*}%
\begin{table*}[htbp]
  \centering
      \scriptsize
      \caption{Comparison of landmark localization results with the state-of-the-art on Bosphorus dataset.}
       \setlength{\tabcolsep}{3.5pt}
       	 \def\pz{\phantom{0}}
       	 \renewcommand{\arraystretch}{1.1}
       \vspace{-3mm}
    \begin{tabular}{|l|l|c|cccccccccccccc|c|}
    \hline
    \multicolumn{18}{|c|}{\textbf{Mean of Localization Error (mm)}} \\
    \hline
    & \multicolumn{1}{c|}{\textbf{Author}} & \textbf{ Images} & \textbf{Ex(L) } & \textbf{Ex( R )} & \textbf{En(L)} & \textbf{En( R)} & \textbf{N} & \textbf{Prn} & \textbf{Ac(L)} & \textbf{Ac ( R)} & \textbf{Ch( L)} & \textbf{Ch( R)} & \textbf{Ls} & \textbf{Li} & \textbf{Pg} & \textbf{Sn} & \textbf{Mean} \\ \hline
    
    \multicolumn{1}{|c|}{\multirow{4}[2]{*}{\rotatebox{90}{Expression}}} 
        & Cruesot et al.~\cite{creusot2013} & 2803  & 6.20  & 4.10  & 5.09  & 4.18  & 6.33  & 4.47  & 4.22  & 4.07  & 4.06  & 8.00  & 7.66  & 5.36  & 8.83  & 15.23 & 6.27 \\
        & Sukno  et al.~\cite{sukno2015} & 2803  & 5.19  & 4.92  & 2.94  & 2.76  & 2.22  & 2.33  & 3.03  & 3.01  & 6.12  & 6.03  & 4.00  & 6.54  & 7.58  & 2.81  & 4.25 \\
        & BFM~\cite{paysan2009}   & 2920  & 3.58  & 3.62  & 2.77  & 2.65  & 2.17  & 2.90  & 3.80  & 4.63  & 5.86  & 6.01  & 3.99  & 6.77  & 8.12  & 3.59  & 4.32 \\
    & \textbf{K3DM$_{BO}$} & 2920  & 3.57  & 4.01  & 2.35  & 2.40  & 2.32  & 2.82  & 2.50  & 2.99  & 4.85  & 4.91  & 3.32  & 5.03  & 6.02  & 2.35  & 3.53 \\ \hline
    
    \multicolumn{1}{|c|}{\multirow{4}[2]{*}{\rotatebox{90}{Rotation}}} 
    & Cruesot et al.~\cite{creusot2013} & 1155  & 5.42  & 4.12  & 5.18  & 3.65  & 5.17  & 4.89  & 3.52  & 3.43  & 4.05  & 4.29  & 3.84  & 3.81  & 4.68  & 9.47  & 4.68 \\
    & Sukno et al.~\cite{sukno2015} & 1155  & 4.48  & 4.95  & 2.97  & 3.23  & 3.40  & 4.36  & 3.36  & 3.37  & 3.76  & 3.75  & 3.47  & 5.01  & 7.77  & 4.19  & 4.15 \\
    & BFM~\cite{paysan2009}   & 1365  & 4.63  & 4.96  & 5.30  & 5.16  & 3.81  & 5.08  & 4.81  & 5.49  & 4.49  & 5.28  & 5.43  & 6.40  & 7.10  & 3.14  & 5.08 \\
    & \textbf{K3DM$_{BO}$} & 1365  & 4.84  & 5.09  & 3.31  & 3.85  & 2.68  & 3.19  & 2.73  & 3.20  & 4.53  & 4.91  & 4.13  & 5.84  & 6.22  & 3.80  & 4.14 \\ \hline
    
    \multicolumn{1}{|c|}{\multirow{4}[2]{*}{\rotatebox{90}{Occlusion}}} 
    & Cruesot et al.~\cite{creusot2013} & 381   & 8.13  & 5.45  & 5.60  & 4.99  & 7.78  & 4.72  & 5.34  & 4.85  & 4.10  & 5.62  & 4.81  & 4.30  & 5.44  & 11.05 & 5.87 \\
    & Sukno  et al.~\cite{sukno2015} & 381   & 6.63  & 6.28  & 3.82  & 3.87  & 4.12  & 3.83  & 4.40  & 4.67  & 4.75  & 5.07  & 3.61  & 4.81  & 7.63  & 3.76  & 4.80 \\
    & BFM~\cite{paysan2009}   & 381   & 4.95  & 4.42  & 3.96  & 3.52  & 2.49  & 3.32  & 4.57  & 4.77  & 3.61  & 3.75  & 3.36  & 4.40  & 5.54  & 2.45  & 3.94 \\
    & \textbf{K3DM$_{BO}$} & 381   & 4.64  & 4.51  & 3.10  & 2.95  & 2.69  & 3.18  & 2.55  & 3.01  & 4.36  & 4.22  & 2.89  & 4.14  & 5.00  & 2.90  & 3.58 \\ \hline
    
    \multicolumn{1}{|c|}{\multirow{6}[2]{*}{All}} 
    & Cruesot et al.~\cite{creusot2013} & 4339  & 6.09  & 4.18  & 5.14  & 4.08  & 6.10  & 4.60  & 4.15  & 3.94  & \textbf{4.05} & 6.83  & 6.37  &\textbf{4.81} & 7.35  & 13.20 & 5.78 \\
    & Sukno  et al.~\cite{sukno2015} & 4339  & 5.13  & 5.05  & \textbf{3.03}  & \textbf{2.98}  & 2.70  & 3.00  & \textbf{3.24}  & \textbf{3.25}  & 5.37  & \textbf{5.34}  & \textbf{3.82}  & 5.98  & 7.63  & 3.26  & 4.27 \\
    & BFM~\cite{paysan2009}   & 4666  & \textbf{3.93} & \textbf{4.03} & 3.41  & 3.34  & 2.68  & 3.57  & 4.07  & 4.86  & 5.34  & 5.65  & 4.37  & 6.50  & 7.64  & 3.38  & 4.48 \\
    & \textbf{K3DM$_{BO}$} & 4666  & \textbf{3.94}  & \textbf{4.15 } & \textbf{2.62} & \textbf{2.80} & \textbf{2.46} & \textbf{2.96} & \textbf{2.55} & \textbf{3.04} & \textbf{4.73}  & \textbf{4.86} & \textbf{3.53} & \textbf{5.21}  & \textbf{6.01} & \textbf{2.75} & \textbf{3.70} \\
    & K3DM$_{BU}$  & 4666  & 4.04  & 4.25  & 3.21  & 3.12  & \textbf{2.50} & 3.27  & 3.65  & 4.34  & 5.16  & 5.45  & 4.25  & 6.25  & \textbf{7.26}  & \textbf{3.16}  & \textbf{4.27} \\
    & K3DM$_{FR}$ & 4666  & 4.13  & 4.27  & 3.33  & 3.24  & 2.60  & 3.51  & 3.91  & 4.61  & 5.26  & 5.56  & 4.32  & 6.43  & 7.48  & 3.26  & 4.42 \\ \hline
    
    \multicolumn{18}{|c|}{\textbf{Standard Deviation of Localization Error (mm)}} \\ \hline
    
     & \textbf{} & \textbf{Images} & \textbf{Ex(L) } & \textbf{Ex( R )} & \textbf{En(L)} & \textbf{En( R)} & \textbf{N} & \textbf{Prn} & \textbf{Ac(L)} & \textbf{Ac ( R)} & \textbf{Ch( L)} & \textbf{Ch( R)} & \textbf{Ls} & \textbf{Li} & \textbf{Pg} & \textbf{Sn} & \textbf{Mean} \\
     \hline
    \multicolumn{1}{|c|}{\multirow{6}[0]{*}{All}} 
    & Cruesot et al.~\cite{creusot2013} &   4339    & 5.02  & 3.79  & 4.43  & 3.49  & 5.22  & 4.61  & 3.45  & 3.11  & \textbf{2.95} & 5.35  & 5.17  & \textbf{3.95} & 8.36  & 10.37 & 4.95 \\
    & Sukno  et al.~\cite{sukno2015} &   4339    & 4.01  & 3.86  &\textbf{2.15}  & \textbf{2.33}  & \textbf{2.27}  & \textbf{2.56}  & 2.37  & \textbf{2.42}  & 5.06  & 4.75  & \textbf{3.51}  & 6.86  & 7.16  & \textbf{2.37} & 3.69 \\
    & BFM~\cite{paysan2009}   &  4666     & \textbf{2.84}  & \textbf{2.97}  & 3.30  & 3.60  & 2.58  & 3.44  & 2.63  & 2.86  & 4.23  & 4.22  & 3.93  & 6.76  & 6.98  & 3.09  & 3.82 \\
    & \textbf{K3DM$_{BO}$ } &  4666     & \textbf{2.69} & \textbf{2.82} & \textbf{2.06} & \textbf{2.23} & \textbf{1.63} & \textbf{1.61} & \textbf{1.45} & \textbf{1.59} & \textbf{3.13}  & \textbf{3.03}  & \textbf{2.99}  & \textbf{4.23}  & \textbf{4.28} & \textbf{2.23} & \textbf{2.57} \\
    & K3DM$_{BU}$ &   4666    & 2.87  & 3.01  & 2.98  & 3.17  & 2.42  & 3.04  & \textbf{2.36}  & 2.56  & 4.01  & \textbf{3.99}  & 3.75  & 6.13  & \textbf{6.20}  & 2.88  & \textbf{3.53} \\
    & K3DM$_{FR}$ &   4666    & 2.93  & 3.07  & 3.10  & 3.30  & 2.52  & 3.26  & 2.54  & 2.72  & 4.10  & 4.07  & 3.80  & 6.31  & 6.39  & 2.97  & 3.65 \\
    \hline
    \end{tabular}%
    \vspace{-3mm}
  \label{tab:BO_LLE_CompC}%
\end{table*}
\begin{table*}[htbp]
\vspace{-0mm}
  \centering
    \scriptsize
    \caption{Comparison of landmark localization results (mean $\pm$ SD) with the state-of-the-art on BU3DFE dataset.}
     \setlength{\tabcolsep}{2.5pt}
     	 \def\pz{\phantom{0}}
     	 \renewcommand{\arraystretch}{1.2}
     \vspace{-2mm}
    \begin{tabular}{|l|c|c|c|c|c|c|c|c|c|c|c|c|}
    \hline
    \textbf{Author} & \textbf{Images} & \textbf{Ex(L) } & \textbf{En(L)} & \textbf{N} & \textbf{Ex( R )} & \textbf{En( R)} & \textbf{Prn} & \textbf{Ac} & \textbf{Ch} & \textbf{Ls} & \textbf{Li} & \textbf{Mean} \\
    \hline
    Nair  et al.~\cite{nair2009}  & 2350  &  {-} &  \multicolumn{1}{l|}{12.1}  &  {-} &  {-} &  \multicolumn{1}{l|}{11.9}  &  \multicolumn{1}{l|}{~8.8}  &  {-} &  {-} &  {-} &  {-} & \multicolumn{1}{l|}{10.9}  \\
    
    Segundo et al.~\cite{segundo2010} & 2500  &  {-} &  {6.3 $\pm$ 4.8} &  {-} &  - &  {6.3 $\pm$ 5.0} &  {1.9 $\pm$ 1.1} &  {6.6 $\pm$ 3.4} &  {-} &  {-} &  {-} &  {4.4 $\pm$ 3.5} \\
    
    Salazar et al.~\cite{salazar2014} & 350   &  {9.6 $\pm$ 6.1} &  {6.8 $\pm$ 4.5} &  {-} &  {8.5 $\pm$ 5.8} &  {6.1 $\pm$ 4.2} &  {5.9 $\pm$ 2.7} &  {6.8 $\pm$ 3.2} &  {-} &  {-} &  {-} &  {5.9 $\pm$ 4.3} \\
    
	Gilani et al.~\cite{gilani2015} & 2500  &  {4.4 $\pm$ 2.7} &  {4.8 $\pm$ 2.6} &  {4.5 $\pm$ 2.7} &  {4.4 $\pm$ 2.7} &  {3.3 $\pm$ 2.7} &  {2.9 $\pm$ 2.0} &  {4.3 $\pm$ 2.7} &  {5.7 $\pm$ 3.7} &  {4.2 $\pm$ 2.7} &  {6.9 $\pm$ 6.3} &  {3.7 $\pm$ 3.1} \\
    
    \textbf{K3DM$_{BU}$} & 2500  &  \textbf{3.8 $\pm$ 2.2} &  \textbf{2.2 $\pm$ 1.5} & \textbf {2.9 $\pm$ 2.1} & \textbf {3.3 $\pm$ 2.2} &  \textbf{2.4 $\pm$ 1.6} & \textbf {2.5 $\pm$ 1.7} & \textbf {2.3 $\pm$ 1.6} &  \textbf{4.6 $\pm$ 3.3} &  \textbf{3.6 $\pm$ 2.3} &  \textbf{6.4 $\pm$ 6.1} & \textbf {2.8 $\pm$ 2.5} \\
    
    K3DM$_{FR}$ & 2500  &  \textbf{4.0 $\pm$ 2.4} & \textbf{2.8 $\pm$ 1.6} &  \textbf{4.3 $\pm$ 2.6} &  \textbf{3.6 $\pm$ 2.4} &  \textbf{2.7 $\pm$ 1.7} &  \textbf{2.6 $\pm$ 1.8} &  \textbf{2.9 $\pm$ 1.8} &  \textbf{5.4 $\pm$ 3.5} &  \textbf{3.8 $\pm$ 2.4} &  \textbf{7.0 $\pm$ 6.6} &  \textbf{3.1 $\pm$ 2.7} \\
    \hline
    \end{tabular}%
    \vspace{-5mm}
  \label{tab:BU_LLE_CompC}%
\end{table*}%
\begin{table}[b]
\vspace{-0mm}
 \centering
       \scriptsize
       \caption{Comparative landmark localisation results (mm) on UND side pose scans.}
        \setlength{\tabcolsep}{1.5pt}
        	 \def\pz{\phantom{0}}
        	 \renewcommand{\arraystretch}{1.1}
        \vspace{-2mm}
    \begin{tabular}{|l|c|c|c|c|}
    \hline
    \textbf{Database} &  {\textbf{DB45L}} &  {\textbf{DB45R}} &  {\textbf{DB60L}} &  {\textbf{DB60R}} \\
    \hline
    
    Yaw Est~\cite{perakis2013} &  {$-45^{\circ}\pm9^{\circ}$} & {$44^{\circ} \pm 8^{\circ}$} & {$-59^{\circ} \pm 8^{\circ}$} & {$57^{\circ} \pm 7^{\circ}$} \\ \hline
    
    \# Scans &  {118} &  {118} & {87} & {87} \\
    \hline
    
    Passalis et al.~\cite{passalis2011} & 6.02 $\pm$ 2.45	& 5.83 $\pm$	2.49	& 6.08 $\pm$	2.53 &		5.87 $\pm$	2.4 \\
    
    Perakis et al.~\cite{perakis2013} & 4.75 $\pm$	1.91 & 5.03 $\pm$1.92	& 	5.30 $\pm$2.49 &	4.95 $\pm$	1.80 \\
    K3DM$_{FR}$  & 4.04 $\pm$	1.77 &	4.31 $\pm$	1.90 &	4.36 $\pm$	2.25 &	4.24 $\pm$	1.28 \\
    \hline
    \end{tabular}%
  \label{tab:EarDB_LLE_Comp}%
\end{table}%

\label{sect:LLE_FRGC}
\textbf{FRGCv2 Dataset}:  We construct a dense correspondence K3DM from the first neutral scan of the first 200 identities (100 males and females each) of this dataset. The remaining 1,956 scans of 266 identities are used as test data. Next, we construct a K3DM from the neutral scans of the next 200 identities (100 male and female each) and use the 2,051 scans corresponding to the first training set for testing. This way, we are able to perform landmark detection on all 4,007 scans of FRGCv2, each time ensuring that the identity used for making the K3DM is not present in the test data.  

We  establish dense correspondences between 9,309 vertices on the FRGCv2 dataset (K3DM$_{FR}$) and report the mean and SD of the Landmark Localization Error ($\epsilon_L$) on 14 fiducial points considered to be biologically significant~\cite{gilani2013}. These anthropometric landmarks are annotated only on the mean face and transferred to each densely corresponded scan in the dataset. Manual annotations provided by Szeptycki et al.~\cite{szeptycki2009} and Creusot et al.~\cite{creusot2013} were used as ground truth for comparison.

A comparison of the mean and SD of landmark localization error of our proposed algorithm with the state-of-the-art in Table~\ref{tab:FRGC_LLE_CompC} shows that our results outperform them by a significant margin. K3DM$_{BU}$ was constructed from 100 neutral expression scans and 100 angry expression level-1 scans. K3DM$_{BO}$ was constructed from the first neutral scan of 105 identities. {The K3DM$_{FR}$ achieves the best performance and even the cross domain K3DMs and the K3DM-NICP variant outperform existing state-of-the-art.} Cumulative localization error plots using K3DM$_{FR}$ are shown graphically in Figure~\ref{fig:LLE_All}(b).

\textbf{{Bosphorus Dataset}}: {We construct two  K3DM$_{BO}$ (100 faces each) from the neutral scans of the Bosphorus dataset~\cite{savran2008} such that the model and test identities are mutually exclusive. Note that there are 299 neutral expression scans in the dataset. We {manually} annotate $14$ fiducial landmarks on the mean face of K3DM and transfer the information to other scans after model fitting. Figure~\ref{fig:LLE_All}(c) shows the cumulative detection rate of the $14$ landmarks. Table~\ref{tab:BO_LLE_CompC} details landmark localization results on the three categories of the Bosphorus dataset. It is evident that our algorithm performs significantly better {than the state-of-the-art} under occlusions, rotation and expression variation. Creusot et al.~\cite{creusot2013} and Sukno et al.~\cite{sukno2015} trained their algorithms on 99 neutral scans. They did not report results on these 99 scans and the scans with yaw rotation of $\pm 90^{\circ}$. On the contrary we report the landmarking results on all 4,666 scans of the database including the scans with large yaw variation. Landmark annotations provided by~\cite{savran2008,creusot2013} were used as ground truth. For this experiment, K3DM$_{FR}$ was created from the neutral scans of first 100 male and female (each) identities of FRGCv2 while K3DM$_{BU}$ was the same as used in experiment on FRGCv2 dataset.}

\textbf{BU3DFE Dataset}: {We construct dense correspondence models from the neutral as well as intensity level-1 anger expression scans of 100 identities of the BU3DFE dataset~\cite{yin2006}. We ensure mutually exclusive test and training identities while landmark localization using a K3DM$_{BU}$. Comparative results on $12$ anthropometric landmarks~\cite{farkas1994} on all 2,500 scans of the dataset are given in Table~\ref{tab:BU_LLE_CompC}. Ground truth landmark locations are provided with the dataset~\cite{yin2006}. Figure~\ref{fig:LLE_All}(d) shows the commutative error detection rate of the $12$ landmarks.} K3DM$_{FR}$ was created from the scans of the first 100 male and 100 female identities of FRGC. Our results are better than the state-of-the-art for both the models.

\newpage
\textbf{{UND Ear Dataset}}: {To evaluate the landmark localization performance of our algorithm on side pose scans containing self occlusions, we perform experiments on the UND Ear Database~\cite{yan2005,yan2006}. We follow the exact protocol outlined by~\cite{passalis2011,perakis2014} for a fair comparison. The dataset is divided into $45^{\circ}$ and $60^{\circ}$ left and right pose scans namely \textit{DB45L}, \textit{DB45R}, \textit{DB60L} and \textit{DB60R}. This is a very challenging dataset due to large yaw rotations, noisy scans and self occlusions. The dense correspondence model is created from $200$ neutral expression scans of FRGCv2. Eight landmarks including the two inner and outer eye corners, nose tip, mouth corners and chin tip are annotated on the mean face of K3DM$_{FR}$.  The mean and SD of landmark localization error for all $8$ points is compared with the state-of-the-art in Table~\ref{tab:EarDB_LLE_Comp}.}

\begin{figure}[tb]

\vspace{-3mm}
\begin{minipage}[b]{0.49\linewidth}
  \centering
\includegraphics[trim = 5pt 0pt 30pt 20pt, clip, width=1\linewidth]{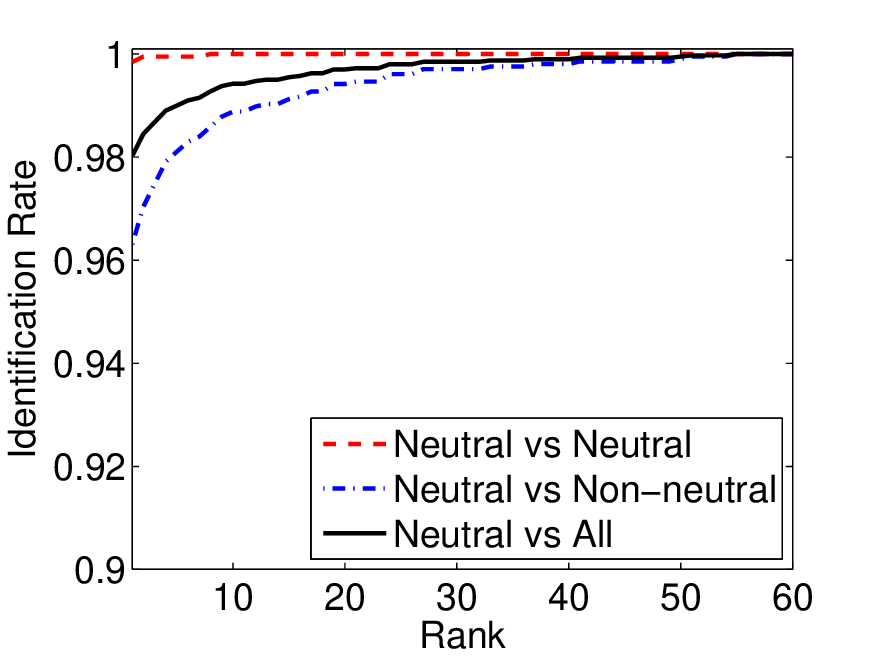}
\end{minipage}
\hfill
\begin{minipage}[b]{.49\linewidth}
  \centering
\includegraphics[trim = 5pt 0pt 30pt 0pt, clip,width=1\linewidth]{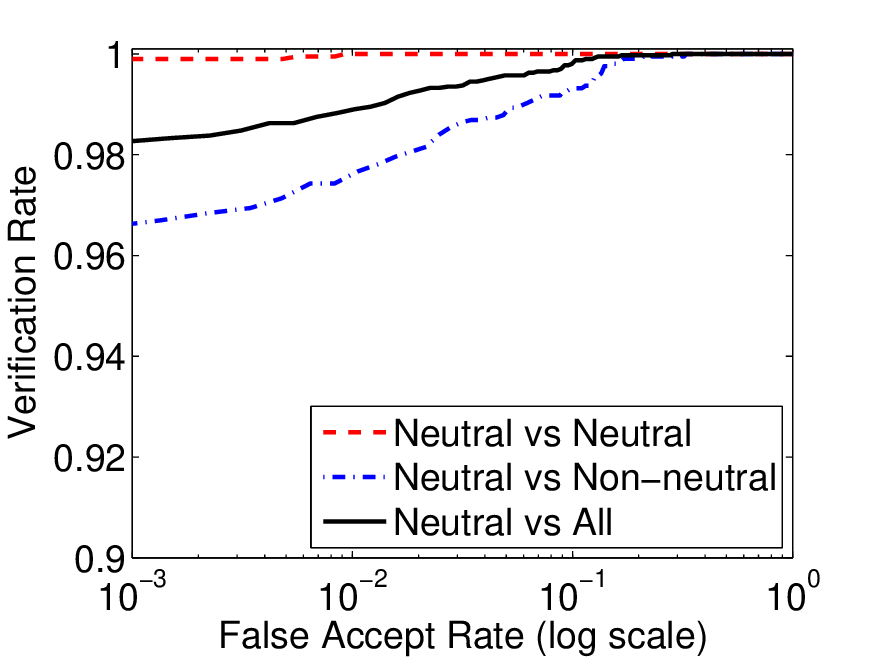}
\end{minipage}
\vspace{-7mm}
\caption{\small{ 
(Left) ROC curves for identification and (Right) verification tasks on FRGCv2 database using our dense correspondence model and fitting algorithms. }}
\label{fig:FaceRegions}
\vspace{-5mm}
\end{figure}

\vspace{-3mm}
\subsection{Face Recognition}
\label{sec:eval-face}
\vspace{-0mm}
{Imaging of faces is considered to be one of the most important biometrics because it can be done passively and is highly distinctive between individuals.} 3D face recognition has addressed many shortcomings of its counterpart in the 2D domain~\cite{mian2007}. We consider this application apt to test the quality of the presented algorithms. Note that our main aim is to evaluate our proposed correspondence and model fitting algorithms as opposed to presenting a face recognition system \textit{per se}. 

\textbf{FRGCv2 Dataset}: We follow the FRGCv2 protocols\cite{phillips2005} of face recognition and include only one scan of each individual (466 in total) in the gallery. To demonstrate the effectiveness of our K3DM Augmentation algorithm, we construct a dense correspondence model on the first available neutral scans of the first 200 identities of FRGCv2 dataset. The constructed model is then augmented with first available neutral scans of the remaining 266 identities of the database using Algorithm~\ref{Algo:DM_Augment}. The probe set consists of the remaining (3,541) scans of all identities. Note that there is only one scan per identity for 56 individuals in the dataset. All of these identities appear only in the gallery. 
The complete dataset was further classified into ``neutral'' and ``non-neutral'' expression subclasses following the protocol outlined in~\cite{mian2008} to evaluate the effects of expressions on deformable model fitting and face recognition.

We employ a holistic and region based approach to model fitting and face recognition. It is well known that the generalization of a model can be increased by dividing faces into independent subregions that are morphed independently~\cite{blanz1999}. This technique has been used extensively for face recognition~\cite{blanz2003,mian2007} and recently for matching off springs to their parents~\cite{dehghan2014}. We also use this approach and perform face recognition by morphing the complete face as well as the eyes and nose regions 
We define these regions on the mean face  which is sufficient to transfer the information to all the faces in the dense correspondence model. 

The full K3DM$_{FR}$ and the eyes and nose models are separately morphed and fitted to each query face in the probe to obtain model parameters ($\alpha$-Step 12 in Algorithm~\ref{Algo:fitting}). Next, the parameters from the whole face and the regions are concatenated to form the feature vector for face recognition. We then perform feature selection using the GEFS algorithm~\cite{gilani2014a,gilani2014b} on the training data set of FRGCv2 containing 953 facial scans. Note that these scans are not used in testing the face recognition algorithm.
The selected features of each query face are matched with those of the gallery faces in the model. The query face is assigned the identity of the gallery face with which it has the smallest distance  $d_f=\cos^{-1} \dfrac{\widetilde{\boldsymbol{\alpha}}_M^T\widetilde{\boldsymbol{\alpha}}_Q}{\|\widetilde{\boldsymbol{\alpha}}_M^T\|_2\|\widetilde{\boldsymbol{\alpha}}_Q\|_2}$, where $\widetilde{\boldsymbol{\alpha}}_M$ are the selected features of each face in K3DM and $\widetilde{\boldsymbol{\alpha}}_Q$ are the selected features of the query face.

\begin{table}[tbp]
  \centering
  \scriptsize
  \caption{Comparison of 3D face recognition results with the state-of-the-art in terms of Rank-1 identification rate (I-Rate) and verification rate (V-Rate) at 0.1\% FAR.}
  \setlength{\tabcolsep}{3.0pt}
  \vspace{-2mm}
    \begin{tabular}{|l|cc|cc|cc|}
    \hline
    \multicolumn{1}{|c|}{\multirow{2}[1]{*}{\textbf{Author}}} & \multicolumn{2}{c|}{\textbf{Neutral}} & \multicolumn{2}{c|}{\textbf{Non-neutral}} & \multicolumn{2}{c|}{\textbf{All}} \\
    \cline{2-7}
	~ & \textbf{I-Rate} & \textbf{V-Rate} & \textbf{I-Rate} & \textbf{V-Rate} & \textbf{I-Rate} & \textbf{V-Rate} \\ \hline
    Mian et al.~\cite{mian2008} 		& \textbf{99.4}\%	& \textbf{99.9}\% & 92.1\% & 96.6\% & 96.1\% & \textbf{98.6}\% \\
    Kakadiaris et al.~\cite{kakadiaris2007} 	& -    	& 99.0\% & -     & 95.6\% & 97.0\% & 97.3\% \\
    Al-Osaimi et al.~\cite{al2009} 	& 97.6\% & 98.4\% & 95.2\% & \textbf{97.8}\% & 96.5\% & 98.1\% \\
    Queirolo et al.~\cite{queirolo2010} 	& -     & 99.5\% & -     & 94.8\% & \textbf{98.4}\% & 96.6\% \\
    Drira et al.~\cite{drira2013} 		& 99.2\% & -     & \textbf{96.8}\% & -     & 97.7\% & 97.1\% \\
    Smeets et al.~\cite{smeets2013}	& - & -     & - & -     & 89.6\% & 79.0\% \\
    Li et al.~\cite{li2014}	& - & -     & - & -     & 96.3\% & - \\
    \textbf{K3DM$_{FR}$}	& \textbf{99.9}\% & \textbf{99.9}\% & \textbf{96.9}\% & \textbf{96.6}\% & \textbf{98.5}\% & \textbf{98.7}\% \\  
	\hline
    \end{tabular}%
  \label{tab:FRGC_FR}%
  \vspace{-6mm}
\end{table}%

\begin{table*}[htbp]
  \centering
   \scriptsize
   \caption{Comparison of Rank-1 recognition results (in \%age) with the state-of-the-art on Bosphorus dataset.}
   \setlength{\tabcolsep}{4.0pt}
    \renewcommand{\arraystretch}{1.2}
   \vspace{-3mm}
    \begin{tabular}{|l|ccc|ccccc|ccccc|c|}
    \hline
    \multicolumn{1}{|c|}{\multirow{3}[2]{*}{\textbf{Author}}} & \multicolumn{3}{c|}{\textbf{Expressions}} & \multicolumn{5}{c|}{\textbf{Poses}}    & \multicolumn{5}{c|}{\textbf{Occlusions}} &  {\multirow{3}[2]{*}{\textbf{All}}} \\
    \cline{2-14}

     & \textbf{AU} & \textbf{Expr} & \textbf{All} & \textbf{YR$<$90} & \textbf{YR90} & \textbf{PR} & \textbf{CR} & \textbf{All} & \textbf{Eye} & \textbf{Mouth} & \textbf{Glasses} & \textbf{Hair} & \textbf{All} &  \\

   {} & 2150  & 647   & 2797  & 525   & 210   & 419   & 211   & 1365  & 105   & 105   & 104   & 67    & 381   & \textbf{4543} \\
    
    \hline
    
    Alyüz  et al.~\cite{alyuz2008} & -     & -     & -     & -     & -     & -     & -     & -     & 93.6  & 93.6  & 97.8  & 89.6  & 93.6  & - \\
    Colombo et al.~\cite{colombo2011} & -     & -     & -     & -     & -     & -     & -     & -     & 91.1  & 74.7  & 94.2  & 90.4  & 87.6  & - \\
    Drira et al.~\cite{drira2013} & -     & -     & -     & -     & -     & -     & -     & -     & 97.1  & 78.0  & 94.2  & 81.0  & 87.0  & - \\
    Berretti et al.~\cite{berretti2013} & -     & -     & 95.7  & 81.6  & 45.7  & 98.3  & 93.4  & 88.6  & -     & -     & -     & -     & 93.2  & 93.4 \\
    Smeetset al.~\cite{smeets2013} & -     & -     & 97.7  & -     & 24.3  & -     & -     & 84.2  & -     & -     & -     & -     & -     & 93.7 \\
    Li et al.~\cite{li2014}	 & \textbf{99.2} & \textbf{96.6}  & \textbf{98.8} & \textbf{84.1}  & \textbf{47.1}  & \textbf{99.5}  & \textbf{99.1}  & \textbf{91.1}  & \textbf{100.0} & \textbf{100.0} & \textbf{100.0} & \textbf{95.5}  & \textbf{99.2} & \textbf{96.6} \\
    
    \textbf{K3DM$_{BO}$} & \textbf{99.0}  & \textbf{96.7} & \textbf{98.5}  & \textbf{99.8} & \textbf{95.2} & \textbf{100.0} & \textbf{99.1} & \textbf{99.0} & \textbf{99.0}  & \textbf{96.1}  & \textbf{100.0} & \textbf{97.3} & \textbf{98.1}  & \textbf{98.6} \\
    \hline
    \multicolumn{15}{l}{AU=Action Units; YR=Yaw Rotation; PR= Pitch Rotation; CR= Cross Rotation}
    \end{tabular}%
  \label{tab:BO_FR}%
  \vspace{-7mm}
\end{table*}%

\begin{table}[htbp]
  \centering
     \scriptsize
     \caption{Comparative of Rank-1 recognition results on partial faces of UND side pose scans.}
     \setlength{\tabcolsep}{4.0pt}
      \renewcommand{\arraystretch}{1.2}
      \vspace{-3mm}
    \begin{tabular}{|l|c|c|c|}
    \hline
    \textbf{Database} & \textbf{UND00LR} & \textbf{UND45LR} & \textbf{UND60LR} \\
    \hline
    \# Scans & {608} & 236   & 174 \\ \hline
    Passalis et al.~\cite{passalis2011} & {76.8\%} & 86.4\% & 81.6\% \\
    Smeets et al.~\cite{smeets2013} & {-} & 98.3\% & 100.0\% \\
    K3DM$_{FR}$ & {86.0\%} & 95.8\% & 98.6\% \\
    \hline
    \end{tabular}%
  \label{tab:EarDB_FR}%
  \vspace{-3mm}
\end{table}%

\begin{table}[htbp]
  \centering
       \caption{Comparison of Rank-1 recognition on FRGCv2 and Bosphorus datasets using cross domain models.}
       \setlength{\tabcolsep}{4.0pt}
        \renewcommand{\arraystretch}{1.2}
        \vspace{-3mm}
    \begin{tabular}{|l|c|c|c|c|c|}
        \hline
        \multicolumn{6}{|c|}{\textbf{FRGCv2}}            \\ 
        \hline
        {\textbf{Method}} & \textbf{Neutral} & \textbf{Expressions} & \textbf{Poses} & \textbf{Occlusions} & \textbf{All} \\ \hline
        BFM~\cite{paysan2009}   & 87.7\% & 65.6\% & -     & -     & 76.4\% \\
        K3DM$_{BU}$ & 92.7\% & 69.0\% & -     & -     & 80.5\% \\
        K3DM$_{BO}$ & 92.1\% & 62.9\% & -     & -     & 77.1\% \\ \hline
        \multicolumn{6}{|c|}{\textbf{Bosphorus}}         \\ \hline
        BFM~\cite{paysan2009}   & -     & 81.1\% & 86.1\% & 86.6\% & 82.7\% \\
        K3DM$_{FR}$ & -     & 85.6\% & 86.5\% & 89.3\% & 85.8\% \\
        K3DM$_{BU}$ & -     & 90.3\% & 92.8\% & 90.7\% & 90.7\% \\
        \hline
        \end{tabular}%
  \label{tab:CD_FR}%
          \vspace{-5mm}
\end{table}%

\begin{figure}[t]
\vspace{+3mm}
\centering
\includegraphics[trim = 0pt 0pt 0pt 0pt, clip, width=.82\linewidth]{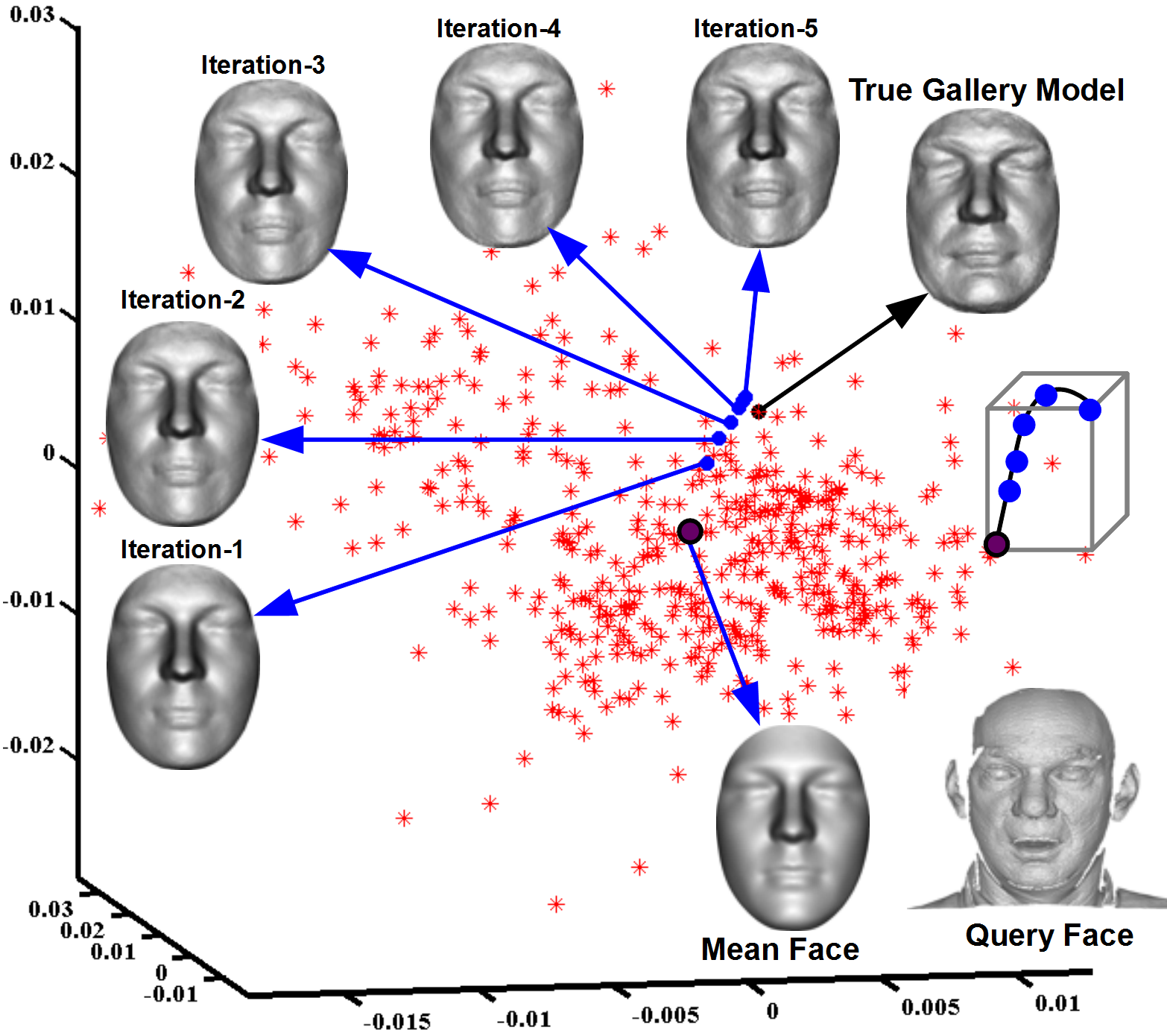}
\vspace{-0mm}
\caption{Iterative model fitting. The 466 FRGCv2 identities are shown as red stars in the first three PC space. The model is morphed iteratively into the query face until the residual error is negligible. Notice how the fitting process takes the query face through a non-linear path (inset image) and removes the extreme facial expression to generate its equivalent neutral expression model.}
\label{fig:FRinPCA}
\vspace{-7mm}
\end{figure}

Figure~\ref{fig:FRinPCA} shows the process of model fitting in PCA space. The dense correspondence model is iteratively fitted on the query face, which in the figure is an  extreme expression scan of the first identity. The model fitting starts from the mean face and in each iteration the fitted query model traverses closer to its gallery face in the PCA space. Face recognition is performed when the fitting residual error $\epsilon_f$ is less than $10^{-5}$.  
Figure~\ref{fig:FaceRegions}(a-b) show the resulting CMC and ROC curves. Rank-1 identification rate for neutral probes is $99.85\%$ while $100\%$ accuracy is achieved at Rank-8. In the more difficult scenario of neutral vs non-neutral, the Rank-1 identification rate is $96.3\%$. A similar trend is observed in the verification rates at 0.1\% FAR. Table~\ref{tab:FRGC_FR} compares our algorithm with the state-of-the-art. In most cases, our results are better than the state-of-the-art depicting the high quality of the dense correspondence model.

\textbf{{Bosphorus Dataset}}: {Experiments are performed on the more versatile Bosphorus dataset to demonstrate the expression, pose and occlusion invariant face recognition capabilities of our proposed model. K3DM$_{BO}$ is formed from the first available neutral scan of each identity in the dataset and a holistic approach to face recognition is adopted. We follow the model fitting and parameter matching technique as mentioned for FRGCv2. Comparative results are given in Table~\ref{tab:BO_FR}. Our proposed technique significantly outperforms the state-of-the-art in pose invariant face recognition, while at the same time it handles expressions and occlusions.}

\textbf{UND Ear Dataset}: {We perform face recognition experiments on this dataset to demonstrate the ability of K3DM to handle pose variations and self occlusions. The dataset is divided into three subsets following the protocol set by~\cite{passalis2011}. \textit{UND00LR} contains 466 subjects of FRGCv2 in the gallery. Two $45^{\circ}$ side scans each (left and right) for 39 subjects and two $60^{\circ}$ side scans each (left and right) for 32 subjects make the probe set. These subjects are common between FRGCv2 and UND Ear databases. \textit{UND45LR} is composed of  $45^{\circ}$ side scans from 118 subjects. The K3DM$_{FR}$ made from $200$ scans is fitted on the left side scan to get the gallery parameters and then fitted to the right side scan to get the probe parameters. A similar protocol is followed for \textit{UND60LR} which contains 60 degree side scans from 87 subjects. Comparative Rank-1 recognition results are given in Table~\ref{tab:EarDB_FR}. Note that while Smeets et al.~\cite{smeets2013} report $>98\%$ face recognition results on the side pose scans of this dataset, their performance on pose variation in the  Bosphorus dataset is significantly low at $84.2\%$. }

\textbf{Cross Domain Face Recognition}: {To compare K3DM with the state-of-the-art Basel Face Model(BFM)~\cite{paysan2009} we perform cross domain face recognition experiments on FRGCv2 and Bosphorus datasets as they include all the challenges of expressions, occlusions and pose variation. For FRGCv2 we use K3DM$_{BO}$ (created from 105 neutral scans of Bosphorus database) and K3DM$_{BU}$ (created from 100 neutral and 100 angry level-1 scans) while for Bosphorus dataset we use K3DM$_{FR}$ and K3DM$_{BU}$. All three models are fitted to each scan in FRGCv2 and Bosphorus datasets. The model parameters of the first neutral scan of each identity in each database are used as gallery features. Table~\ref{tab:CD_FR} details the Rank-1 recognition results from this experiment which show that K3DM outperforms the BFM.} 

\vspace{-3mm}
\section{Discussion and Conclusions}
\vspace{-1mm}

 {We have proposed an algorithm that simultaneously establishes dense correspondences between a large number of 3D faces. Based on the dense correspondences, a deformable face model was constructed. We also proposed morphable model fitting and updating algorithms that are useful for landmark identification and face recognition. Thorough experiments were performed on synthetic and real 3D faces. Comparison with existing state-of-the-art shows that our algorithm consistently achieves better or comparable performance on both the tasks on all datasets. It is interesting to note that while the face recognition algorithm proposed by Li et al.~\cite{li2014} performs well on Bosphorus database, it does not fare that well on FRGCv2. Similar trend can be observed in case of Smeets et al.~\cite{smeets2013} for face recognition and Sukno et al.~\cite{sukno2015} for landmark localization. To the best of our knowledge this is the first paper that has reported consistent comparable results on a variety of application on four public datasets.}

{Although the dense correspondence model assumes frontal and neutral pose scans, for landmark localization and face recognition it demonstrates robustness to occlusion as well as pose and expression variation during the fitting process. Hence the three proposed algorithms present a unified solution to a variety of applications under expression, occlusion and pose variation. The model can handle pose variation up to $\pm 90^{\circ}$. }

{With regards ot the computational complexity, it may be noted that the model building process has to be done off-line. The algorithm iterates over geodesic patches between vertices for each image. Building a dense correspondence model on $105$ identities of  Bosphorus database in approximately $30$ iterations took over $48$ hours on a Core$^{TM}$--i7 machine with $8$GB RAM using MATLAB$^{TM}$. However, the model fitting process on an unseen face takes less than seven seconds.}


\ifCLASSOPTIONcaptionsoff
  \newpage
\fi

\vspace{-2 mm}
\renewcommand{\baselinestretch}{1}
\bibliographystyle{IEEEtran}
\vspace{-1 mm}


\renewcommand{\baselinestretch}{1}

\vspace{-13 mm}
\begin{IEEEbiography}[{\includegraphics[width=1.0in,height=1.20in,clip,keepaspectratio]{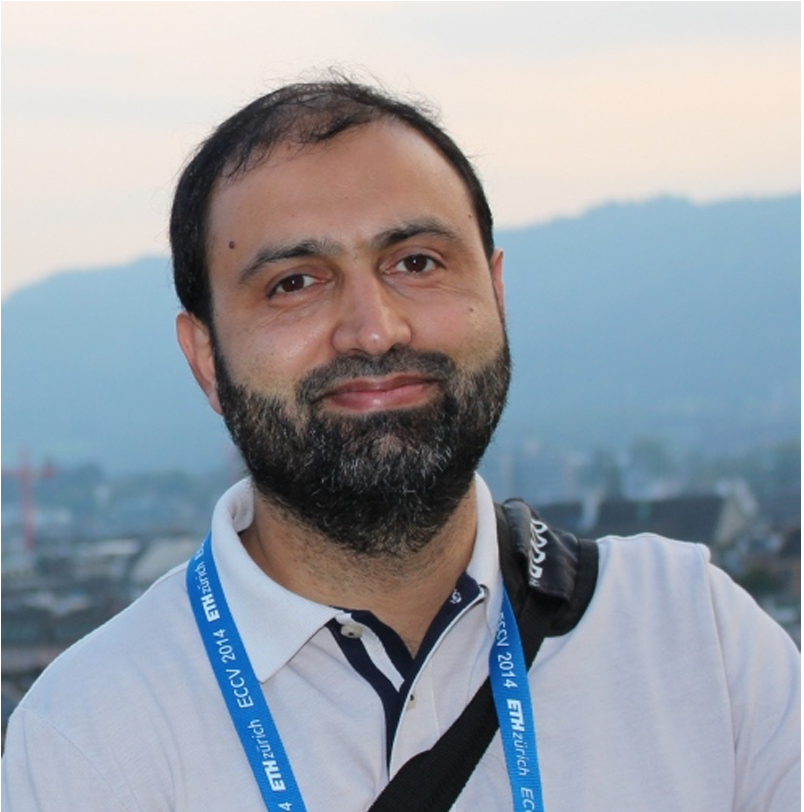}}]{Syed Zulqarnain Gilani}
   did his MS in EE from the National University of Sciences and Technology (NUST), Pakistan in 2009 and secured the Presidents Gold Medal. He recently completed his PhD from the University of Western Australia. His research interests include 3D facial morphometrics with applications to syndrome delineation  and machine learning.
\end{IEEEbiography}
\vspace{-15 mm}

\begin{IEEEbiography}[{\includegraphics[width=1.0in,height=1.20in,clip,keepaspectratio]{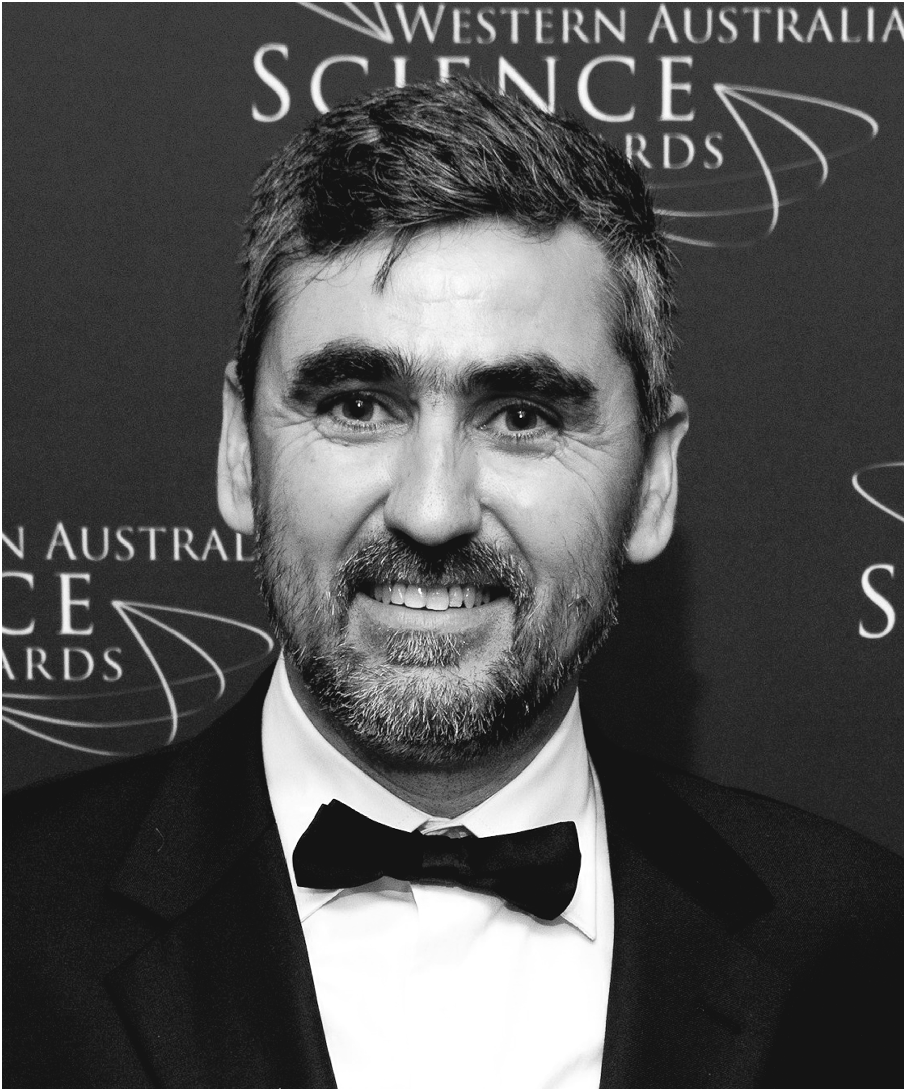}}]{Ajmal Mian}
  is an Associate Professor of Computer Science at The University of Western Australia. He has received several awards including the West Australian Early Career Scientist of the Year Award, the Vice-chancellors Mid-career Research Award and the Outstanding Young Investigator Award. He has received two prestigious fellowships and seven major grants 
  with total funding of \$3.0 Million. 
\end{IEEEbiography}
\vspace{-14 mm}

\begin{IEEEbiography}[{\includegraphics[width=1.10in,height=1.20in,clip,keepaspectratio]{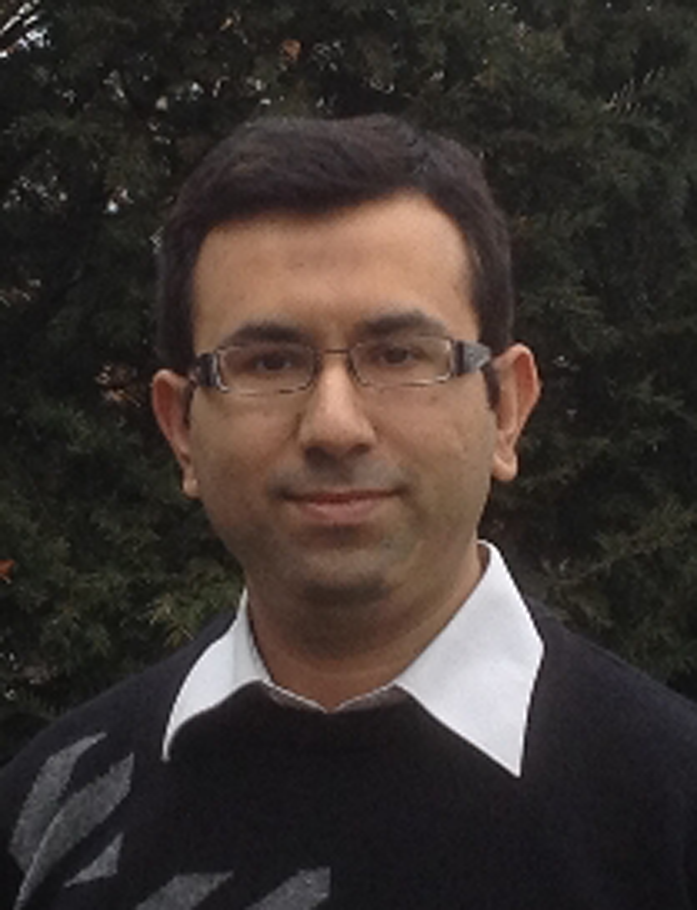}}]{Faisal Shafait}
   is working as an Assistant Professor at The University of Western Australia. Formerly, he was a Senior Researcher at the German Research Center for Artificial Intelligence (DFKI) Germany.  He received his Ph.D. in computer engineering from Kaiserslautern University of Technology, Germany in 2008. 
\end{IEEEbiography}
\vspace{-14 mm}

\begin{IEEEbiography}[{\includegraphics[width=1.10in,height=1.20in,clip,keepaspectratio]{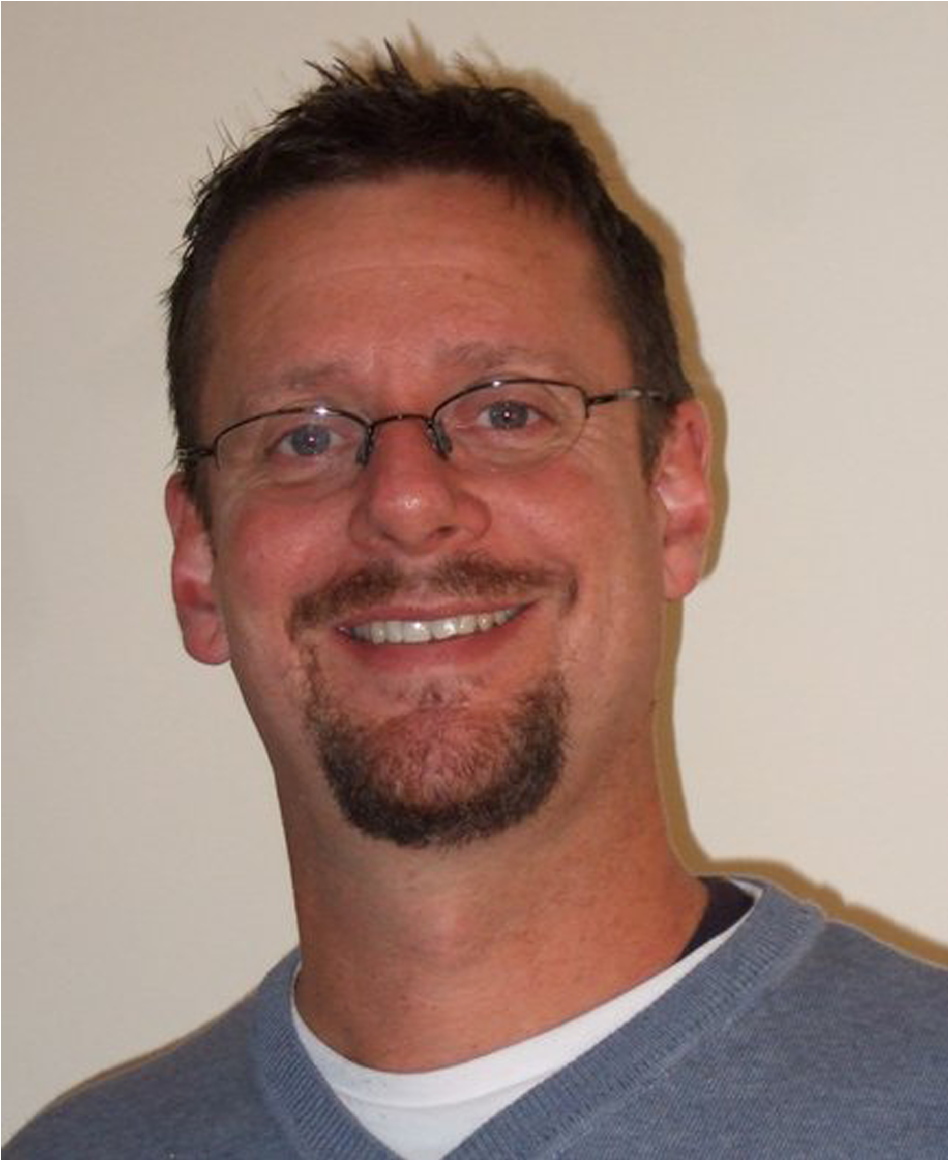}}]{Ian Reid}
   is a professor of Computer Science at the University of Adelaide. He received the DPhil degree from University of Oxford in 1991.  He has been  employed in the Robotics Research Group, conducting research in computer vision, including holding an EPSRC Advanced Research Fellowship (1997-2000), and he has been a University lecturer since 2000. In 2005, he was awarded the title of reader and in 2010 the title of professor. 
\end{IEEEbiography}
\vspace{-5 mm}

%
%
%

\end{document}